%% file: main.tex
\def\year{2021}\relax
\documentclass[letterpaper]{article} % DO NOT CHANGE THIS
\usepackage{aaai21}  % DO NOT CHANGE THIS
\usepackage{times}  % DO NOT CHANGE THIS
\usepackage{helvet} % DO NOT CHANGE THIS
\usepackage{courier}  % DO NOT CHANGE THIS
\usepackage[hyphens]{url}  % DO NOT CHANGE THIS
\usepackage{graphicx} % DO NOT CHANGE THIS
\usepackage{natbib}  % DO NOT CHANGE THIS AND DO NOT ADD ANY OPTIONS TO IT
\usepackage{caption} % DO NOT CHANGE THIS AND DO NOT ADD ANY OPTIONS TO IT
\usepackage{mathtools, nccmath}
\usepackage{multirow}
\usepackage{amssymb}% http://ctan.org/pkg/amssymb
\usepackage{pifont}% http://ctan.org/pkg/pifont
\newcommand{\xmark}{\ding{55}}%
\usepackage{graphicx}
\usepackage[ruled,vlined, linesnumbered]{algorithm2e}
\usepackage{url}
\usepackage{amsmath}
\usepackage{xcolor}
\usepackage{subfigure}
\usepackage{hyperref}% DO NOT CHANGE THIS
\title{Relaxed Clustered Hawkes Process for Procrastination Modeling in MOOCs}
\author {
        Mengfan Yao,\textsuperscript{\rm 1}
        Siqian Zhao, \textsuperscript{\rm 1}
        Shaghayegh Sahebi, \textsuperscript{\rm 1} 
        Reza Feyzi Behnagh \textsuperscript{\rm 2} \\
}
\affiliations {
    \textsuperscript{\rm 1} Department of Computer Science,
        University at Albany - SUNY\\
    \textsuperscript{\rm 2} Department of Educational Theory \& Practice, University at Albany - SUNY\\
    myao@albany.edu, szhao2@albany.edu,  ssahebi@albany.edu,rfeyzibehnagh@albany.edu
}

\begin{document}

\maketitle

\begin{abstract}
Hawkes processes have been shown to be efficient in modeling bursty
sequences in a variety of applications, such as finance and social network activity analysis.
Traditionally, these models parameterize each process independently and assume that
the history of each point process can be fully observed.
Such models could however be inefficient or even prohibited in certain real-world
applications, such as in the field of education, where such assumptions are violated.    
Motivated by the problem of detecting and predicting student procrastination in students Massive Open Online Courses (MOOCs) with missing and partially observed data, in this work, we propose a novel personalized Hawkes process model (\textit{RCHawkes-Gamma}) that 
% predicts future activities with fewer parameters to be inferred.  
% To do this, our model 
discovers meaningful student behavior clusters by jointly learning all partially observed processes simultaneously, without relying on auxiliary features.
Our experiments on both synthetic and real-world education datasets show that RCHawkes-Gamma can effectively recover student clusters and their temporal procrastination dynamics, resulting in better predictive performance of future student activities.
Our further analyses of the learned parameters and their association with student delays show that the discovered student clusters unveil meaningful representations of various procrastination behaviors in students.

\end{abstract}
\section{Introduction}
\input{intro}

\section{Related Work}
\label{sec:related}
\input{related-new.tex}

\section{Problem Formulation}
\label{sec:problem}
\input{problem}
\section{Model: Relaxed Clustered Hawkes}
\label{sec:model}
 
\input{model}

\section{Experiments}
\label{sec:exp}
In this section we evaluate our approach with several state-of-the-art competitors on both simulated and real datasets. 

\noindent\textbf{Setup.} In simulated data, we randomly select a ratio of $r = [0.1,0.3,0.5,0.7]$ amount of students' last two assignment activities to be entirely missing (unseen set), and for the rest of the student-assignment pairs, the first $70\%$ of activities are used for training (training set) and the last $30\%$ are used for testing (seen set). 
In both real datasets, the unit time is set to be an hour, and we use activities that took place before the mid point of the last assignment as training. Hyperparameters of the proposed and baseline models are tuned via grid search.

\subsection{Baselines}
\input{baseline}

\subsection{Datasets}
\input{dataset}

% \subsection{Experiment Setup}
% \input{metrics}
% \label{sec:metrics}

\subsection{Fit and Prediction Performance of RCHawkes-Gamma}
\label{sec:results}
\input{results}

\subsection{Student Procrastination in RCHawkes-Gamma}
\label{sec:analysis}
\input{analysis}

\section{Conclusion}
\label{sec:conclusion}
\input{conclusion}
\section{Acknowledgement}
This paper is based upon work supported by the National
Science Foundation under Grant Number 1917949.
% \section{Appendix}
\input{algorithm}
\input{intuition}

\bibliography{refs.bib}

% \newpage
% \section*{Appendix}
% \subsection*{Appendix A}
% \input{AuthorKit21/supplementary/appendix}
% \newpage
\end{document}

%% file: intro.tex
%---------------------------------------- topic + terminology
% ----- procrastination is bad, especially in online
Academic procrastination, or postponing the starting of planned studies, has been associated with negative side-effects on students' academic performance, psychological well-being, and health~\citep{Moon2005,Steel2007}.
This behavior is more prevalent in online educational settings, that require high levels of time-management and self-regulation skills~\cite{Lee2011} and can lead to low academic outcomes and course drop-outs~\cite{Vitiello2018}.
%---------------------------------------- motivation: why important
% ----- it can be used in finding ways to reduce procrastination by nudging students
With the growth of online education,
%and employers' trust in it, online academic 
%procrastination behavior can be costly, both for the learners and for the employers~\cite{Grunschel2013,Radford2014}.
%Thus, 
it is essential to devise mechanisms to detect the potential future procrastination in students, to be able to prevent this behavior and its associated negative consequences.

% ----- procrastination can lead to bursty interaction
%Time-management and self-regulation are negatively associated with academic procrastination~\cite{Zimmerman2008,Azevedo2010}.
In studies on self-reported academic procrastination, %this behavior has been linked with the lack of time-management and self-regulation skills~\cite{Zimmerman2008,Azevedo2010}.
%As a result, 
this behavior is indicated by cramming of studying activities: given a time interval followed by a deadline, students show limited studying activities at the beginning of the interval, followed by a burst of studying (cramming) closer to the deadline~\citet{Perrin2011,Gelman2016}.
However, these studies do not provide a unified quantitative definition of procrastination, other than qualitative student self-reports, that can be scarce and hard to obtain.
% \textcolor{red}{However, these studies do not provide a unified quantitative definition of procrastination, other than qualitative student self-reports. However self-reports could be unreliable, since they rely on students' memory of their previous behavior, rather than how they actually study.}
Prior work also shows that although each student has their individual studying habits, students can be clustered into a few distinct groups by their studying behaviors~\citep{Yao2020,uzir2020analytics}.
In essence, in highly procrastinating students, getting closer to the deadline may trigger more intense studying activities, while in others, their studies are more regulated and distributed across the time interval.
%---------------------------------------- problem
% Despite these findings, in most of the few studies on student procrastination, the temporal aspects of students' behavior have been ignored~\cite{Cerezo2017,Kazerouni2017}, or were not personalized~\cite{Baker2016,Park2018,Backhage2017}.
Despite these finding, most of the studies on student procrastination either ignored the temporal aspects of students' behavior~\cite{Cerezo2017,Kazerouni2017}, or were not personalized for students~\cite{Baker2016,Park2018,Backhage2017}.
%or ~\cit{Backhage2017}.
% Backhage et al. proposed a model that captures procrastination-deadline cycles of all students in the course using a stochastic temporal model~\cite{Backhage2017}. % dynamics of student population interaction with online courses 
% However, this model assumes that all students follow the same procrastination behavior during the course and does not distinguish the differences between student behaviors.
% Park et al. model student procrastination using a mixture model of per-day student activity counts during each week of the course~\cite{Park2018}.
% Each day of the week is associated with a Poisson rate parameter that is constant during the whole course.
% Despite representing individual student activity counts, this model cannot differentiate between different weeks in the course, does not have a continuous time scale, and cannot capture non-homogeneously spaced deadlines in a course.
More importantly, current research cannot predict when student's next activity will take place.
Ideally, a procrastination model can capture the underlying \textit{cluster structures} in student activity sequences, can be \textit{personalized} to capture different students' studying habits, and can deal with \textit{unseen data} such as assignments that are not yet started by students, and represent students' activity burstiness.

% ----- we want to model procrastination's burstiness for students and predict their next delay + burstiness
We note that Hawkes processes~\cite{Hawkes1971} have the potential to represent students' procrastination behavior, as they model activity burstiness, as opposed to memoryless Poisson processes. 
%Particularly uni-variate Hawkes processes can be a good representative of student sequences as neither a students' activity triggers another student's activity, nor the deadline of one assignment triggers the activities that are related to another assignment.   
%They have been successfully used in modeling self-exciting temporal events in various application domains, including social networks~\cite{Zhao2015,Chen2019} and recommender systems~\cite{Du2015,Xiao2016,Jing2017}.  
% Especially, because students' activity sequences do not affect or excite each other, they can be modeled using different uni-variate Hawkes processes.
% However, the vanilla uni-variate Hawkes process is not personalized, if all activity sequences are modeled with the same parameter sets, or has to learn many parameters, if each activity sequence is modeled with a different parameter set.
% Additionally, 
However, when modeling one sequence per user-item pair, conventional Hawkes processes model each item's sequences individually and do not rely on the similarities between different items.
Thus, they cannot infer parameters for items that have unseen data~\cite{hosseini2016hnp3,choi2015constructing,mei2017neural,Du2016}.
% assume that the whole activity sequences are observed and thus, cannot infer unseen data~\cite{}.
In some recent work, low-rank personalized Hawkes models aim to address this problem ~\cite{Du2015}, usually with the help of auxiliary features to reinforce the low-rank assumption~\cite{Shang2018,shang2019geometric}.
Yet, to the best of our knowledge, none of the previous Hawkes models were able to represent the cluster structure between sequences, while being personalized and inferring unseen data.
%traditional models of uni-variate Hawkes processes do not perfectly fit our problem as they assume that the whole activity sequences are observed.
%Even in some of the newer research that are developed to infer unseen data, the .
%they miss some aspects of an ideal procrastination model 
%they miss at least one aspect of an ideal procrastination model: some do not represent the underlying cluster structure of the sequences, some are not personalized, and some cannot deal with the unseen data. 

%However, driven by our application domain, we focus on uni-variate Hawkes processes.

%Typically, an event sequence modeled by Hawkes process is represented by parameters such as base intensity and burstiness.
%Having many event sequences, recent research has focused on reducing the number of parameters to be estimated, by capturing the potential structures in the intensity and burstiness parameter matrices. %by imposing constraints such as rank or 

%In this paper, we propose a novel Relaxed Clustered Hawkes process with a Gamma prior (\textit{RCHawkes-Gamma}) to model and predict the cramming behavior, as a proxy to procrastination, in students of online courses.
In this paper, we propose a novel Relaxed Clustered Hawkes process with a Gamma prior (\textit{RCHawkes-Gamma}) to model and predict the cramming procrastination behavior in students of Massive Open Online Courses (MOOCs).
% In our setting, all deadline-related student activities are triggered by the deadline and the activity sequences of different deadlines do not directly affect each other.
% \textcolor{blue}{Our proposed model captures individual students' cramming behaviors while discovering clusters of students with similar behaviors, and infers missing student activities both on unseen and unfinished assignments.}
To do this, we model each student-assignment pair, that is the interactions of a student with a course assignment characterized by activity times, as a uni-variate Hawkes process. 
% By capturing the group structures among students, our proposed model can predict students' activities in future assignments, as a representation of procrastination, based on only partially observed historical activities. 
By modeling all student-assignment pairs jointly, our proposed model is able to capture similarities shared among students (i.e. cluster structures) by learning a low-dimensional representation of procrastination (i.e. personalization).
As a result, even for student-assignment pairs without observed history (i.e. unseen data), their parameters can be inferred based on the group structure, without relying on auxiliary features or historical observations.

% Putting together the intensity and burstiness parameters of each student-assignment pair, we build parameter matrices for each of these measures.
% Assuming similarity structures between student behaviors, we impose a clustering-structure regularization on the student dimension of these parameter matrices, in addition to the low-rank regularization on the assignment dimension.
% Learning these parameters, in addition to discovering the similarity structures between different processes, we can predict students' activities in future assignments, as a representation of procrastination, without observing their historical data on those particular assignments. 
% RCHawkes-Gamma can be used in other application domains, such as recommending recurrent items to users.

More specifically, our contributions are:
(1) We propose a Relaxed Clustered Hawkes model, driven by the problem of modeling academic procrastination in MOOCs;
(2) Our \textit{personalized} model represents the \textit{similarity structure} between multiple event sequences without requiring auxiliary features (Section~\ref{sec:model}) and infers \textit{unseen data} in event sequences;
(3) We experiment on both synthetic and real-world datasets to show that the proposed model can recover clusters of students and their temporal procrastination dynamics, resulting in a better predictive performance of future activities (Section~\ref{sec:exp}); and
(4) We further study the learned parameters to demonstrate that the discovered student clusters are meaningful representations of various procrastination-related behaviors in students (Section~\ref{sec:analysis}).

%---------------------------------------- why it's hard?
% in EDM: because we have deadlines and it's reversed timeline
% because students may exhibit group/cluster behavior, but they do not affect each other
% and assignments do not affect each other too
% excitement by deadline

%---------------------------------------- what's been done?
% low-dimensional multivariate/mukltidimensional hawkes has been done before.

%---------------------------------------- the gap
% ----- why we're not using multidimensional hawkes? 
% multidimensional is good for *social* because it captures the influence *between* individuals (mutually-exciting property)
% but we don't have this in education

% ----- why we're not using other low-rank one-dimensional hawkes? 
% baseline? no auxiliary information? only in the student dimension?

%---------------------------------------- research questions

%---------------------------------------- key components of approach and results + limitations

%---------------------------------------- summary of contributions w. falsifiable claims

%% file: related-new.tex
\textbf{Low-Rank Hawkes Processes}
\label{sec:related:hawkes}
Hawkes processes~\cite{Hawkes1971} have been successfully used %to model processes 
in applications such as social networks%~\cite{Zhou2013,Linderman2014,Zhao2015, Farajtabar2017,
~\citep[e.g.][]{Chen2019}, mobility patterns~\citep[e.g.][]{Vassoy2019}, %,Wang2017,Vassoy2019},Du2016 
and recommender systems~\citep[e.g.][]{Du2015}. %Xiao2016.
%For example, Du et al. proposed a topic-sensitive information diffusion model in social networks based on multivariate Hawkes processes to estimate the topic-dependent transmission rates~\cite{Du2013}.
%However, their application in the field of Educational Data Mining is limited.
%\noindent\textbf{Low-Rank Hawkes Processes}
Among them, the most relevant work to ours is low-rank uni-variate Hawkes processes proposed by Du et al., to capture user-item interaction and recommend items ``at the right moment''~\cite{Du2015}. 
%Their experiments include recommending items to users according to a ranked list of estimated item intensities for each user and predicting the returning time for an interaction between a user and an item by average sampling from the density of the next event time, both evaluated by mean absolute error.
However, this work does not incorporate the clustering behavior that is essential in our problem domain.
% We include this model as one of our baselines.
%The most relevant literature on Hawkes processes to our work is low-rank or structured Hawkes processes inspired by reducing the number of learned parameters~\cite{Zhou2013,Du2015,Shang2019}.
%Other than the work by Du et al.~\cite{Du2015}, 

Other relevant literature on Hawkes processes mainly falls into 3 categories: 
(1) Multi-variate Hawkes processes that focus on modeling the mutual excitation among sequences~\cite{Zhou2013,Luo2015,Bacry2015,Lemonnier2017}. 
% For example, in modeling news propagation trends across different social networks, one can assume that, activities from different social network platforms can trigger each other (e.g. tweets trigger Facebook posts).
% an activity in one social network (e.g., a tweet in twitter) can trigger an activity in another social network (e.g., a post in Facebook). 
% In our application of modeling students' procrastination behavior however, neither a students' activity triggers another student's activity, nor the deadline of one assignment triggers the activities that are related to another assignment.
% Since the assignments are not parallel, activities for each assignment are only triggered by the same assignment's deadline.
% Also, since students do not work in teams and are not in a social setting, studying activities of each student can only be the offspring of the same student's past activities.
% Hence, we build our model based on the uni-variate Hawkes process.
(2) Uni-variate Hawkes models that model each sequence \textit{independently} and discard the potential relatedness among all sequences, thus cannot infer sequence's future when its history is not observed 
% the arrival times of the future activities without observed history, 
e.g.~\cite{mei2017neural,du2016recurrent,Du2015a,xiao2017modeling,Li2018,li2020tweedie}. 
For example, Du et al. propose to use RNN to model the arrival times of a given sequence to capture more complicated sequence dynamics compared to traditional Hawkes models~\cite{du2016recurrent}. 
Such RNN-based models predict future time after time $t$ based on the observed history unfolded up to time $t$, therefore cannot directly predict sequences that do not have historical observations;
(3) Approaches that \textit{jointly} model different sequences as uni-variate Hawkes processes by capturing the similarities among the sequences (e.g. via a low-rank constraint). Therefore, they can predict the future events for the sequences without historical observations, by utilizing histories from sequences that are structurally similar.
However, such methods usually rely on auxiliary information~\cite{He2015,Li2018,Shang2018,shang2019geometric}. 
For example, in the recommender system setting, Shang et al. impose a local low-rank constraint on the parameter matrix to model large-scale user-item interactions by first computing user-item pairs' similarities via item features~\cite{Shang2018}.
% using auxiliary information, such as item features, that capture the similarity between user-item pairs~\cite{Shang2018}.
In contrast, due to privacy constraints in our application, many educational datasets are highly anonymized and scarce.
Consequently, having a model that does not require such information is valuable in our context.

\noindent
\textbf{Procrastination Modeling in Education Domain}
As there is no quantitative definition for procrastination behavior, in most of the recent educational data mining literature, procrastination-related behavior has been summarized by curating time-related features from student interactions in the course.
These studies aim to evaluate the relationships between these time-related features with student performance and do not model temporal aspects of procrastination~\cite{Baker2016,Cerezo2017,Kazerouni2017,agnihotriprocrastination}.
The few recent works that model student activity sequences, are limited in their assumptions, do not capture student activity intensities, are not personalized, do not model time dependencies between student actions, and do not infer missing data~\cite{Park2018,Yao2020}. 
%Capturing the temporal aspects of procrastination, such as activity intensity or burstiness, has been attempted in a few numbers of recent works~\cite{Backhage2017,Park2018}.
For example, Park et al. classify students to procrastinators vs. non-procrastinators by formulating a measure using a mixture model of per-day student activity counts during each week of the course~\cite{Park2018}.
But, it cannot model non-homogeneously spaced deadlines in a course.
Furtheremore, even though each student's activity is counted in a daily basis, it is not a temporal approach that models activity time points. 
%It assumes that students' activity \textit{counts} can be modeled by a mixture Poisson distribution, thus they are able to compute the likelihood of students having more activities towards the end of the week or not. 
Indeed, none of these models can predict \textit{when} the future activities will happen. % and cannot model non-homogeneously spaced deadlines in a course.
Similarly, Backhage et al. proposed Poisson distribution to model students' daily activity count in order to capture procrastination-deadline cycles of all students in the course~\cite{Backhage2017}. 
% dynamics of student population interaction with online courses 
% However, this model assumes that all students follow the same procrastination behavior during the course and does not distinguish the differences between student behaviors.
In their work, each day of the week is associated with a Poisson rate parameter that is constant during the whole course.
Despite representing individual student activity counts, this model cannot differentiate between different weeks in the course, does not have a continuous time scale, and cannot capture non-homogeneously spaced deadlines in a course.

To the best of our knowledge, the only model that can be compared to ours in predicting activity times %uses Hawkes processes to model student procrastination behavior 
is a Hawkes process model by Yao et al.~\cite{Yao2020} that relates procrastination to the mutual excitation among activity types.
% uses activity types as auxiliary information, to model student procrastination behavior.
This work does not model student behavior clusters, and cannot infer unseen data.
%Since this model is the closest research on procrastination modeling to our work,
We use this model, called EdMPH, as one of our baselines.
% , even though it uses a multi-variate Hawkes model.

%% file: problem.tex
Our goal is to model partially observed student-assignment interactions and predict two types of future student-assignment interactions: 1) future assignments with no historical activities (\textit{unseen data requirement}), and 2) current assignments that students are working on (assignment-student pairs with \textit{partially observed} history).

Specifically, we consider a course that includes $N$ non-parallel assignments and $M$ students.
Each student $u_j$ can perform a sequence of activities towards each assignment $a_i$, such that each sequence is indexed by a student-assignment pair $(a_i,u_j)$.
Activities in a sequence are presented with a timestamp that marks their arrival time.
% ($x_{i,\tau}^j$ for the $\tau$-th activity of student $u_j$ on assignment $a_i$). 
We assume that the activities within each student-assignment pair happen either because they are a pre-requisite for another assignment-related activity (\textit{internal stimuli}), or because of a non-activity related reason (\textit{external stimuli}).
For an example of internal stimuli, think of when students divide their big tasks (e.g., submitting the final assignment response) into smaller sub-tasks (e.g., solving a sub-problem of the assignment), within each sub-task, one activity spontaneously leads to another related activity. 
Conversely, external stimuli can come from the individual student's tendency to study regularly or due to the assignment deadline~\footnote{As student activities are triggered by the upcoming deadlines from the future but not the past, without loss of generalizability, we use a reversed activity timeline for our data.}.
On the other hand, we assume no causal relationship between student-assignment pairs: since assignments are not parallel, activities towards assignments do not trigger each other. 
Further, since students do not work in teams and are not in a social setting, there are no triggering effects between student activities.
We also assume that while students having their individual learning pattern towards each assignment (\textit{personalization assumption}), their studying activities follow a latent structure that can group students with similar learning behaviors (\textit{cluster assumption}).

%% file: model.tex
According to our problem formulation and assumptions, we build our model based on uni-variate Hawkes processes.
The reason behind our choice of the model is two-fold: (1) Unlike the memoryless Poisson process that assumes the independence among activities, Hawkes can model the aforementioned internal and external stimuli that exist in student activities; (2) Unlike the multi-variate Hawkes processes that assume triggering effects between dimensions, there are no exciting effects between assignments or student sequences.
%since our goal is not to model possible excitation between activity types or among student-pairs
We first present the intensity function that defines student-assignment pairs. %describe by uni-variate Hawkes processes; 
We then add low-rank and relaxed clustering constraints to capture our personalization and cluster assumptions, and add a Gamma prior to address the unseen data requirement. 
\subsection{Uni-Variate Hawkes Intensity Function} 
% Since our goal is to study the sequences of student activities and their inter-arrival times, point processes are of the best choices for our application.
% Unlike the memoryless nature of the Poisson process that assumes the independence between past and future activities , we need a type of point process that can handle the event dependency resulted from student activities' internal stimuli.
% Therefore, we build our model based on the Hawkes process that expects the activities to be excited both externally (similar to the Poisson process) and internally, i.e. self-excited offspring activities.
%As mentioned in Section~\ref{sec:related:hawkes}, 
% Uni-variate Hawkes processes are the best fit for our application as the event dependencies in our application happens within each sequence and not between different sequences.
% We further add low-rank and relaxed clustering constraints to the model to capture our personalization and cluster assumptions, and a Gamma prior to address the unseen data requirement. 

Formally, given a sequence of activities for an assignment-student pair $(a_i,u_j)$, we model its activities' arrival times $X_i^j = \{x_{i\tau}^j|\tau = 1,...,n_i^j\}$ by a uni-variate Hawkes process, via the intensity function of time $t$, defined as follows~\cite{Hawkes1971}:
% ~\footnote{
% % Note that in regular applications of Hawkes, an activity at time $t$ can trigger later activities at times $\tau > t$. 
% % However, in our application, student activities are triggered by the upcoming deadlines in the future and sub-tasks at times $\tau < t$ can be offsprings of future studying tasks at time $t$.
% % So, 
% We use a reversed activity timeline for our data as student activities are triggered by the upcoming deadlines from the future but not the past.
% This does not affect our model, optimization, or learned parameters.}:
%\vspace{-0.1in}

% \begin{equation}
%     \lambda(t) = \mu + \sum_{x_{\tau} < t} \alpha \beta \exp(-\beta (x_{\tau} - t)),
%     % \label{eq:basic}
%     %\vspace{-0.1in}
%     \label{eq:unionehawkes}
% \end{equation}

% where $x_{\tau}$ represents the time of $\tau-$th activity that is observed before time $t$, $\alpha$ models the expected number of offspring activities that are directly related to the previous ones (the self-exciting or bursty nature of the process); $\mu$ is the average arriving rate of activities in the long run (i.e. base rate); and $\beta$ represents how fast the historical activities will stop affecting the future activities (i.e. decay). 

% \begin{equation}
%      \Lambda(t) = U + A\beta\sum_{\tau = 1}^{n_{ij}}   \exp(-\beta (t - \mathcal{X}_\tau)),
%      \label{eq:matrix}
% \end{equation}
\vspace{-10pt}
\footnotesize{\begin{equation}
     \lambda(t)_{ij} = U_{ij} + A_{ij}\beta\sum_{\tau = 1}^{n_{ij}}   \exp(-\beta (t - x^j_{i,\tau})),
     \label{eq:matrix}
     \vspace{-5pt}
\end{equation}}\
where $x^j_{i,\tau}$ is the $\tau$-th element in the vector $X^j_i\in \mathbb{R}^{n_{ij}}$, which denotes the arrival time of the $\tau$-th activity that belongs to assignment-student pair $(a_i,u_j)$, $n_{ij}$ is the total number of observed activities for $(a_i,u_j)$;
%Suppose there are $N$ assignments and $M$ students,
$U\in \mathbb{R}^{N\times M}$ is the non-negative base rate matrix, where $U_{ij}$ quantifies the expected number of activities that are triggered externally within $(a_i,u_j)$;
$A\in \mathbb{R}^{N\times M}$ is the non-negative self-excitement matrix, with $A_{ij}$ representing the self-exciting or bursty nature of $(a_i,u_j)$, i.e., the expected number of activities that are triggered by the past activities; and $\beta$ is a global decay rate that represents how fast the historical activities stop affecting the future activities. 

\subsection{Relaxed Clustered Hawkes}
Conventional uni-variate Hawkes processes model each process individually.
% , meaning that no assumptions are made on the collection of students (i.e. columns of $A$ and $U$) or assignments (i.e. rows of $A$ and $U$). 
In this work, we assume that the occurrences of assignment activities and their characteristics, parameterized by Hawkes process, are similar among some students, but less similar to some others, i.e. parameter matrix $A$ exhibits cluster structure on its columns. 
% In other words, we assume that the self-excitement matrix $A$ exhibits cluster structure among students, where the similarity within a cluster is stronger than between cluster similarities.

Particularly, we assume that students form $k<M$ clusters %in which each student $i$ is described by 
according to their behaviors towards all assignments represented in $A$'s column vectors. % of $(\alpha_{1i},...,\alpha_{Ni})$ in $A$. 
To impose this, we add a clustering constraint to our model using the sum of squared error (SSE) penalty, similar to K-means clustering: %\cite{ding2004k} 
%as follows:
\footnotesize{\begin{align}
\label{eq:kmeans_org}
    P(A,W) & = \rho_1 \text{tr}(A^\top A - W^\top A^\top AW) + \rho_2 \text{tr}(A^\top A)\\\nonumber
    &=\text{tr}(A((1+\frac{\rho_1}{\rho_2})I - WW^T)A^\top,
\end{align}}\
    \vspace{-10pt}

where $\rho_1$ and $\rho_2$ are regularization coefficients; $W\in\mathbb{R}^{M\times k}$ is an orthogonal cluster indicator matrix, with $W_{ij} = \frac{1}{\sqrt{n_j}}$ if $i$ is in $j$-th cluster, and 0 otherwise (showing which students belong to which cluster); and $n_j$ is the size of cluster $j$. 

Since this strict constraint is non-convex, we follow Jacob et al.'s work~\cite{jacob2009clustered} to obtain its convex relaxation problem:
\footnotesize{\begin{align}
\label{eq:kmean}
     \min&\mathcal{L}_{c}(A,Z) = \min \frac{\rho_2(\rho_2+\rho_1)}{\rho_1} tr(A (\frac{\rho_1}{\rho_2} I + Z)^{-1} A^\top)\\\nonumber
   &\text{s.t. } tr(Z) = k, Z\preceq I, Z\in S_{+}^M.
\end{align}}\
$Z = WW^\top\in\mathbb{R}^{M\times M}$ represents cluster-based similarity of students, with $W$ defined in Eq.~\ref{eq:kmeans_org}. 
Here, the trace norm is a surrogate of the original assumption that there are $k$ clusters and the other two constraints are the relaxation of $W$ being orthogonal. As a result, this equation is jointly convex to both $A$ and $Z$. 
We call this model \textit{RCHawkes}.

\subsection{A mixture Gamma prior}
% When inferring Hawkes parameters, a common way is to use Maximum Likelihood Estimate. However, 
% To improve our model's ability to infer Hawkes parameters, especially in the case of unseen data, we use a mixture Gamma prior on the self-excitement matrix $A$. 
To improve our model's robustness to potential outliers and to possibly reduce overfitting, we add a mixture Gamma prior on the self-excitement matrix $A$.
As a result, the summation of the first three terms in Equation~\ref{eq:cost} is the A-Posteriori estimation, which not only is more robust comparing with Maximum Likelihood Estimation, also it provides an interpretation of each component's hyperparameters in student clusters: i.e. the pseudo counts of externally and internally excited activities. 
% To improve our model's ability to infer Hawkes parameters, especially in the case of unseen data, we use a mixture Gamma prior on the self-excitement matrix $A$ to achieve our proposed \textit{RCHawkes-Gamma} model. 
Specifically, consider the prior for $A_{ij}$ when student $i$ is in $m$-th cluster:
\footnotesize{
\begin{align}
    p(A_{ij}; \Theta_m) &= \frac{1}{\Gamma(s_m)\theta_m^{s_m}}A_{ij}^{s_m-1}\exp(-\frac{A_{ij}}{\theta_m^{s_m}}),
\end{align}}\
where $\Theta_m = (s_{m},k_m)$, are hyperparameters which respectively control the shape and the scale of the gamma distribution in cluster $m$. 
%We call this model \textit{RCHawkes-Gamma}.
The loss brought by the mixture Gamma prior can be computed as follows:
\footnotesize{
\begin{align}
\label{eq:Lg}
    &\mathcal{L}_{g} = \log p(A; \Theta_1,..,\Theta_k)\\\nonumber
    %  & = \mathcal{L} + \sum_{j}\sum_{i} \log [p(\alpha_{ji}\sim Gamma(\alpha_{ji}; \Theta_m))\\
    %  &p(\alpha_{ji}| \alpha_{ji}\sim Gamma(\alpha_{ji})]\\
    & =  \sum_{X_i^j\in \mathcal{O}} \big[\log \sum_{m =1}^k  \frac{1}{k}\frac{1}{\Gamma(s_m)\theta_m^{s_m}}A_{ij}^{s_m-1}\exp(-\frac{A_{ij}}{\theta_m^{k_m}})\big],
\end{align}}\
where $\mathcal{O}$ is the collection of all observed $X_i^j$.
% \textcolor{orange}{Adding this mixture Gamma prior not only }
% The explicit form of this mixture Gamma prior is given in , where $s_{m}$ and $\theta_{m}$ are hyperparameters which respectively control the shape and the scale of the $m$-th Gamma distribution that describes cluster $m$.
% %\noindent\textbf{
\subsection{Objective Function}
For our model, we need to consider the multiple sequences (as in Eq.~\ref{eq:matrix}) and add the introduced constraints. 
Here we first introduce a recursive function $R$ and matrix $T$ that can be computed offline to ease the computation. 
% The motivation to introduce the recursive function $R(\tau)$ is to avoid the double summation term in Eq.~\ref{eq:matrix}. 
% Specifically, we have: 
% \vspace{-10pt}
{\footnotesize
\begin{align}
    R_{ij}(\tau)  = 
\begin{cases} 
\big(1+R_{ij}(\tau -1)\big)\exp(-\beta\big(x^j_{i,\tau}-x^j_{i,\tau-1})\big) & \text{if $\tau>1$,} \\
0 & \text{if $\tau =1$}.
\end{cases}
\end{align}
}\
We also construct the matrix $T$ as follows to avoid repetitive computation in iterations:
{\footnotesize
\begin{align}
T = [\sum_{\tau=1}^{n_{ij}}( \exp(-\beta(x^j_{i,n_{ij}} - x^j_{i,\tau}))-1)]_{N\times M}
\end{align}
\vspace{-10pt}
}\

To this end, the final objective function of our proposed model, given the observed activities for all assignment-student pairs $X$ can be described as in Eq.~\ref{eq:cost}. 
{\footnotesize
\begin{align}
\label{eq:cost}
   &\min_{A\geq0,U\geq0,Z}  -L(X;A,U)  \\\nonumber
   &= -\sum_{X_i^j\in\mathcal{O}}\sum_{\tau = 1}^{n_{ij} }\log \big (U_{ij} + A_{ij}\beta R^j_{i}(\tau)\big) + U_{ij} x^j_{i,n_{ij}}\\\nonumber
   &+ A\circ T -\mathcal{L}_{g}(A; \Theta_1,..,\Theta_k) + \mathcal{L}_c(A,Z)+\rho_3tr(A)\\\nonumber
   &\text{s.t. }% tr(Z) = k, Z\preceq I, Z\in S_{+}^M, 
   A\geq 0, U\geq 0,
\end{align}
}\
where $\mathcal{L}_c$and $\mathcal{L}_g$ are the previously defined losses introduced by clustering and gamma prior respectively and $\rho_3$ is a regularization coefficient.
The trace norm regularization, is a convex surrogate for computing rank of $A$, which enables the knowledge transfer from the processes with observations to the unseen assignment-user pairs that do not have any observed historical activities. Finally, %the constraints on $Z$ are first introduced in Equation~\ref{eq:kmean} that capture clustering structure, and 
to not violate the definition of Hawkes process, we have non-negative constraints on $A$ and $U$.
% Finally, the last term first introduced in Eq.~\ref{eq:kmean} is the the convex relaxation of the K-means clustering objective, which assumes column-wise cluster structure, and $k$ is the parameter that encodes our assumption in terms of the number of clusters observed among students.

\subsection{Optimization}
To solve the minimization problem in Eq.~\ref{eq:cost}, we could use the Stochastic Gradient Descent algorithms.
However, the non-negative constraints on $A$ and $U$ along with the non-smoothed trace norms can complicate the optimization.
To tackle this problem, we used the Accelerated Gradient Method~\cite{nesterov2013gradient}. 
The key component of using this method is to compute the following  proximal operator:
\footnotesize{
\begin{align}
\label{eq:proximal}
     &\min_{A_z,U_z,Z_z} \|A_z-A_s\|_F^2 + \|U_z-U_s\|_F^2 + \|Z_z-Z_s\|_F^2 + \\\nonumber
& \text{s.t. } tr(Z_z) = k, tr(A_z) \leq c, A_z\geq0, U_z\geq0, Z_z\preceq I, Z_z\in S_{+}^M\nonumber 
% \label{eq:proximal}
\end{align} }\
where subscripts $z$ and $s$ respectively represents the corresponding parameter value at the current iteration and search point~\cite{nesterov2013gradient}. 
We present Algorithm~\ref{algo} to efficiently solve the objective function using the Accelerated Gradient Descent framework. ~\footnote{Details of the algorithm, its complexity, convergence analyses, and our code can be found in \url{https://github.com/persai-lab/AAAI2020-RCHawkes-Gamma}.}
% Details of the algorithm, its complexity, and convergence analyses are in the supplementary material.
% \vspace{-10pt}
\input{algo}

%% file: algo.tex
%\vspace{-10pt}
\begin{algorithm}[h]
% \tiny
\footnotesize
\SetAlgoLined
\SetKwInput{KwOutput}{Output}  
\KwIn{$\eta > 1$, step size $\gamma_0$, $\rho_{3}$, MaxIter}
initialization: $A_1 = A_0; U_1 = U_0; Z_1 = \frac{k}{M} \times I$; $\alpha_{0} = 0; \alpha_1 = 1$\;
\For{$i = 1$ to MaxIter}{
  $a_i = \frac{\alpha_{i - 1} -1}{\alpha_{i}}$\;
  $S_i^A = A_i + a_i(A_i - A_{i - 1})$\;
  $S_i^B = U_i + a_i(U_i - U_{i - 1})$\;
  $S_i^Z = Z_i + a_i(Z_i - Z_{i - 1})$\;
     \While{Ture}{
      Compute $A_* = \mathcal{M}_{S^A_i,\gamma_i}(A)$ \\
      = $\big (\text{TrPro}(S_i^A-\nabla \mathcal{L}(A)/\gamma_i, \rho_{3})\big )_+$ \;
      Compute $U_*  = \mathcal{M}_{S^U_i,\gamma_i}(U)$ \;
      Eigen-decompose $S_i^Z = Q\Sigma Q^{-1}$\;
      Compute $\underset{\sigma^*_i}{\operatorname{argmin}} \sum \limits_{i} (\sigma_i - \hat{\sigma}_i)^2$, $\sum \limits_{i} ^M \sigma_i  = k,~0\leq \sigma_i\leq 1$\;
      Compute $\Sigma_{*} = diag(\sigma^*_1,...,\sigma^*_M)$\;
      Compute $Z_* = Q\Sigma_{*} Q^{-1}$\;
      \eIf{$\mathcal{L}(A_*, U_*, Z_*) \leq \mathcal{L}(S_i^A, S_i^U, Z_i) + \sum_{x\in\{A,U,Z\}}\langle S_i^x,\delta \mathcal{L}(S_i^x)\rangle +  \alpha_k/2\|S_i^x - x_*\|_F^2$}{
       break\;
       }{
       $\gamma_i = \gamma_{i-1} \times \eta$\;
      }
     
  $A_{i+1} = A_*$; $U_{i+1} = U_*$; $Z_{i+1} = Z_*$\;
  \eIf{stopping criterion satisfied}
  {break\;}
  {$\alpha_i = \frac{1 + \sqrt{1 + 4 \alpha_{i-1}^2}}{2}$}
     }
     }
\KwOutput{$A = A_{i+1}, U = U_{i+1}, Z = Z_{i + 1}$}
 \caption{Accelerated PGA}
 \label{algo}
\end{algorithm}

%% file: baseline.tex
%To evaluate the performance of the proposed model (RCHawkes-Gamma), 
We consider two sets of state-of-the-art baselines: the ones that are able to infer unseen data, and the ones that cannot. 
A summary of all baseline approaches is presented in Table~\ref{tbl:baselines}.
In the following we briefly introduce each of the baselines.

\noindent\textbf{EdMPH}: A Hawkes model that was recently proposed to model student procrastination in Educational Data Mining domain~\cite{Yao2020}. It applies a Multivariate Hawkes Model which utilizes student activity types as extra information, and cannot infer unseen data.

\noindent\textbf{RMTPP}: A Recurrent Neural Network Hawkes model to represent user-item interactions~\cite{du2016recurrent}. It does not directly infer parameters of unseen data and it uses activity markers (i.e. features) as an input.

\noindent\textbf{ERPP}: A similar approach to baseline RMTPP, but it includes time-series loss in the loss function~\cite{xiao2017modeling}.

\noindent\textbf{HRPF} and \textbf{DHPR}: Two Poisson factorization models proposed in \cite{hosseini2018recurrent} that do not require user-network as auxiliary features. These models, however, do not directly model the time-dependencies between the future and the past, thus cannot quantify activity self-excitement.

\noindent\textbf{HPLR}: An item recommendation model using Hawkes process \cite{Du2015}. It is the most similar to ours, as it imposes a low rank assumption on matrices $A$ and $U$ and can infer unseen data.
However, unlike our model, it does not consider the cluster structure of parameter matrix $A$.

\noindent\textbf{RCHawkes}: A variation of our proposed model that does not use a Gamma prior. Its objective is to find the maximum of log-likelihood rather than the maximum of A-posterior.

\begin{table}[]
\caption{A summary of baseline approaches.}
\vspace{-10pt}
\resizebox{0.45\textwidth}{!}{
\begin{tabular}{|c|c|c|c|c|}
\hline
Application & Model &\begin{tabular}[c]{@{}c@{}}Infer\\ Unseen\\Data\end{tabular} & \begin{tabular}[c]{@{}c@{}}Require No\\ External\\ Features\end{tabular} & \begin{tabular}[c]{@{}c@{}}Model Time \\Dependency\end{tabular} \\ \hline
 & RCHawkes-Gamma & \checkmark & \checkmark & \checkmark \\ \cline{2-5} 
 & RCHakwes & \checkmark & \checkmark & \checkmark \\ \cline{2-5} 
\multirow{-3}{*}{EDM} & EdMHP & \xmark & \xmark & \checkmark \\ \hline
 & HPLR & \checkmark & \checkmark & \checkmark \\ \cline{2-5} 
 & ERPP & \xmark & \xmark & \checkmark \\ \cline{2-5} 
 & HRPF & \checkmark & \checkmark & \xmark \\ \cline{2-5} 
 & DRPF & \checkmark & \checkmark & \xmark \\ \cline{2-5} 
\multirow{-5}{*}{Rec-Sys} & RMTPP & \xmark & \xmark & \checkmark \\ \hline
\end{tabular}}
% \caption{A summary of baseline approaches and their categories.}
\label{tbl:baselines}
\vspace{-10pt}
\end{table}

%% file: dataset.tex
%We evaluate our model on a synthetic and two real-world datasets as described in the following.

\noindent
\textbf{Synthetic Data}
To create simulated student-assignment pairs, we first construct the parameter matrices. 
We build $A_s$ by: 
%To build $A_s$: 
a) sampling $k=3$ sets of column $\alpha$'s from different Gamma distributions, for different student procrastination behavior clusters; 
b) adding white noise ($\sigma^2=0.1$); and % to both $A_s$ and $U_s$; and
c) shuffling all columns randomly to break the order.
%We used Gamma distribution for matrix $A_s$ since it is a common choice for describing rate distribution as it is conjugate with the exponential family distributions. 
%As we do not have a similar assumption on the base rate matrix 
We build $U_s$, by sampling it from a normal distribution.  %we sample each matrix element from a normal distribution. 
Then, we sample $150$ activities for each assignment-student pair using the Ogata thinning algorithm~\cite{ogata1988statistical}, which is the most commonly used sampling method in the related literature. 
Finally, we obtain $5400$ simulated student-assignment pairs and $810$K synthetic activities.  
% We also two online courses from two MOOC platforms that contain historical time stamps of students activities.

% We also experiment on two online courses from two MOOC platforms that allow us to get the students' history activities.

% from Canvas Network~\footnote{http://canvas.net} and Coursera\footnote{https://educational-technology-collective.github.io/morf/platform/}. 

\noindent
\textbf{Computer Science Course on Canvas Network (CANVAS)} 
%The first online dataset that we consider 
This real-world MOOC dataset is from the Canvas Network platform~\cite{Canvas-Network2016}. 
Canvas Network is an online platform that hosts various open courses in different academic disciplines. 
%The courses offered on this platform have multiple types of learning resources, which includes learning modules, assignments/quizzes and discussions. 
%The course we select is in the computer science discipline. 
%The history of student activities, such as the time that a student submits a quiz or participates in a discussion thread are included. 
The computer science course we use happens during $\sim6$ weeks. 
%The duration of this computer science course is about $6$ weeks. 
In each week, an assignment-style quiz is published in the course resulting in $6$ course assignments. 
In total, we extract $\sim740$K assignment-related activity timestamps from $471$ students.
%, which includes submission activities, module learning activities, and discussions that are also associated with the $6$ course assignments.

\noindent
\textbf{Big Data in Education on Coursera (MORF)} Our second real-world dataset % course we consider 
is collected from an 8-week ``Big Data in Education'' course in Coursera platform. 
The dataset is available through the MOOC Replication Framework (MORF)~\cite{andres2016replicating}. 
% It is a online platform include data of Coursera courses. The course we select in our works is named as Big Data in Education. 
%This course is 8-week long. 
%Within each week, it provides students with multiple types of learning resources such as assignments, video lectures, and discussion forums.
%It contains students' interaction histories with learning material.%, for instance, the timestamps that students submit a quiz, watch a video, or post in a discussion forum. 
In total, we extract $\sim102$K activities of $675$ students related to $8$ assignments. 
%In the rest of this paper, we refer these two datasets as CANVAS and MORF datasets.

%% file: results.tex
\subsubsection{Estimated Parameters on Simulated Data}
\label{sec:res:paramest}
In the simulated dataset, as we know the true parameters (i.e. $A$ and $U$), we compare the Root Mean Squared Error (RMSE) of estimated $\hat{A}$ and $\hat{U}$, varying unseen data ratio $r$~\footnote{Baselines ERPP, RTMPP, and EdHawkes cannot be used in this analysis, since they parameterize the processes differently.}. 
% since the intensity functions in baseline approaches ERPP, RTMPP and EdHawkes do not follow the definition of the uni-variate Hawkes process's intensity function (i.e. a base rate term plus a memory kernel function), we can not compare the estimated Hawkes parameters and the correlation matrix of $\hat{A}$ with these models. 
The results %of estimated $A$ and $U$ 
are presented in Tbl.~\ref{tbl:syn_a}. 
RCHawkes-Gamma and RCHakwes outperform the baseline methods usually by a large margin, for both the sequences with seen and unseen history.
Also, even though all models perform worse with the increase of $r$, RCHawkes-Gamma and RCHakwes' RMSEs have a lower standard deviation, indicating less variation in their performance even in high missing data ratios. 
% We also notice that, even though the performances of all models decrease with the increase of missing ratio, prediction MSEs of RCHawkes-Gamma and RCHakwes usually have a much lower standard deviation, indicating less variation of the performance in learning the parameters when the number of missing data is high. 
%It is also worth mentioning that the performances of the models in modeling unseen data are generally worse than their performances in modeling the processes with observed historical activities. 
Additionally, the models' performances in unseen data are generally worse than their performances in the processes with observed historical activities. 
One possible reason is that the Hawkes parameters for unseen data in this simulation can only be inferred from the similar processes with observed data by leveraging the row and column relatedness, while the true characteristics of the unseen processes can not be entirely captured as there are no observations that the models can use.

\begin{table*}[]
\centering
\vspace{-15pt}
\caption{RMSE ($\pm$standard deviation) of $\hat{A}$ and $\hat{U}$ on seen and unseen data, with various missing data ratios ($r$)}
   \vspace{-10pt}
\label{tbl:syn_a}
\resizebox{0.98\textwidth}{!}{
\begin{tabular}{|c|c|c|c|c|c|c|c|c|c|}
\hline
\multirow{7}{*}{\begin{tabular}[c]{@{}c@{}}RMSE\\ for A\end{tabular}} & \multirow{2}{*}{Model} & \multicolumn{2}{c|}{r = 0.1}                                                                                               & \multicolumn{2}{c|}{r = 0.3}                                                                                               & \multicolumn{2}{c|}{r = 0.5}                                                                                               & \multicolumn{2}{c|}{r = 0.7}                                                                                               \\ \cline{3-10}
                        && seen & unseen &  seen & unseen & seen & unseen & seen & unseen\\  \cline{2-10}
&RCHawkes-Gamma         & \textbf{0.094$\pm$0.024 }                                               & \textbf{0.102$\pm$0.037}                                                  & 0.121$\pm$0.017                                                & \textbf{0.114$\pm$0.056}                                                  & 0.141$\pm$0.033                                                & 0.139$\pm$0.033                                                  & \textbf{0.136$\pm$0.077}                                                & \textbf{0.137$\pm$0.052}                                                  \\
&RCHawkes              & 0.108$\pm$0.017                                                & 0.108$\pm$0.054                                                  & \textbf{0.115$\pm$0.024}                                                & 0.116$\pm$0.039                                                  & \textbf{0.126$\pm$0.033}                                                & \textbf{0.136$\pm$0.033}                                                  & 0.180$\pm$0.072                                                 & 0.170$\pm$0.048                                                   \\
&HPLR                   & 0.631$\pm$0.110                                                & 0.663$\pm$0.331                                                  & 0.645$\pm$0.141                                                & 0.607$\pm$0.216                                                  & 0.635$\pm$0.133                                                & 0.633$\pm$0.133                                                  & 0.634$\pm$0.304                                                & 0.634$\pm$0.204                                                  \\
&HRPF                   & 0.664$\pm$0.769                                                & 0.664$\pm$0.769                                                  & 0.664$\pm$0.770                                                & 0.664$\pm$0.770                                                  & 0.663$\pm$0.769                                                & 0.663$\pm$0.770                                                  & 0.664$\pm$0.769                                                & 0.664$\pm$0.767                                                  \\
&DRPF                  & 0.474$\pm$0.461                                                & 0.474$\pm$0.461                                                  & 0.479$\pm$0.465                                                & 0.479$\pm$0.465                                                  & 0.473$\pm$0.462                                                & 0.473$\pm$0.462                                                  & 0.474$\pm$0.463                                                & 0.474$\pm$0.463               \\
\hline

\multirow{5}{*}{\begin{tabular}[c]{@{}c@{}}RMSE\\ for $U$\end{tabular}} & 
RCHawkes-Gamma         & 0.075$\pm$0.022                                                                    & 0.085$\pm$0.036                                                                      & \textbf{0.069$\pm$0.017 }                                                                   & \textbf{0.060$\pm$0.050 }                                                                      & \textbf{0.062$\pm$0.030 }                                                                   & \textbf{0.064$\pm$0.030 }                                                                     & 0.071$\pm$0.039                                                                    & 0.075$\pm$0.026                                                                      \\
&RCHawkes               & 0.074$\pm$0.020                                                                     & 0.089$\pm$0.061                                                                      & 0.074$\pm$0.020                                                                    & 0.075$\pm$0.032                                                                      & 0.077$\pm$0.030                                                                    & 0.079$\pm$0.030                                                                      & \textbf{0.069$\pm$0.026 }                                                                   & \textbf{0.062$\pm$0.017}                                                                     \\
&HPLR                   & 0.110$\pm$0.082                                                                     & \textbf{0.078$\pm$0.047}                                                                      & 0.081$\pm$0.060                                                                    & 0.078$\pm$0.094                                                                      & 0.091$\pm$0.035                                                                    & 0.091$\pm$0.035                                                                      & 0.090$\pm$0.096                                                                     & 0.095$\pm$0.065                                                                      \\
&HRPF                   & 0.105$\pm$0.055                                                                    & 0.311$\pm$0.055                                                                      & 0.119$\pm$0.068                                                                    & 0.183$\pm$0.068                                                                      & 0.141$\pm$0.071                                                                    & 0.142$\pm$0.071                                                                      & 0.179$\pm$0.068                                                                    & 0.120$\pm$0.070                                                                      \\
&DRPF                   & \textbf{0.062$\pm$0.052 }                                                                  & 0.300$\pm$0.035                                                                        & 0.088$\pm$0.049                                                                    & 0.165$\pm$0.045                                                                      & 0.121$\pm$0.051                                                                    & 0.121$\pm$0.050                                                                      & 0.167$\pm$0.053                                                                    & 0.102$\pm$0.054   \\ \hline
\end{tabular}}
\vspace{-10pt}
\end{table*}

\begin{figure*}[ht]

\centering

\subfigure{
\includegraphics[width=0.545\textwidth]{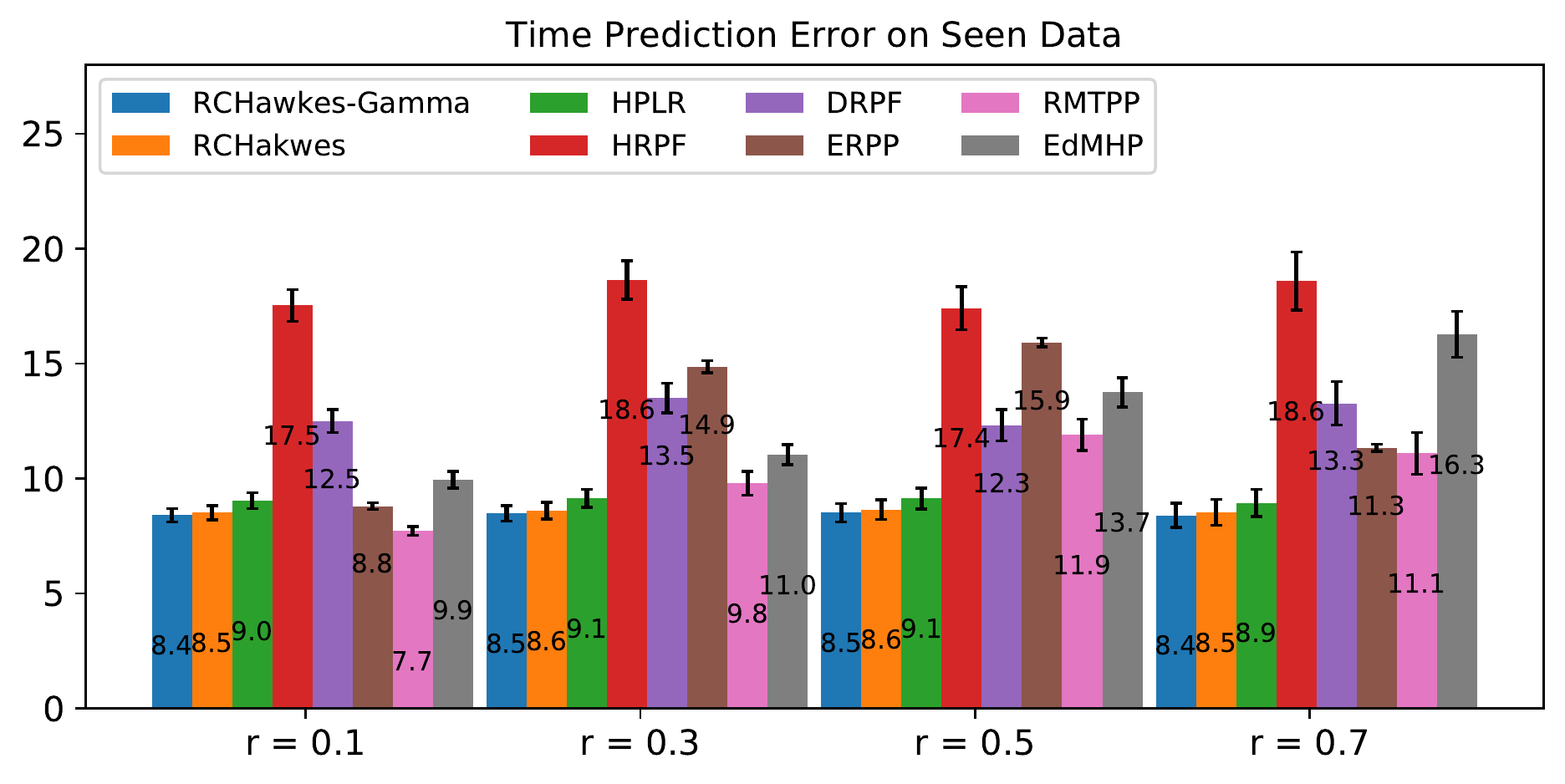}
}
 \hspace{-10pt}
\subfigure{
\includegraphics[width=0.435\textwidth]{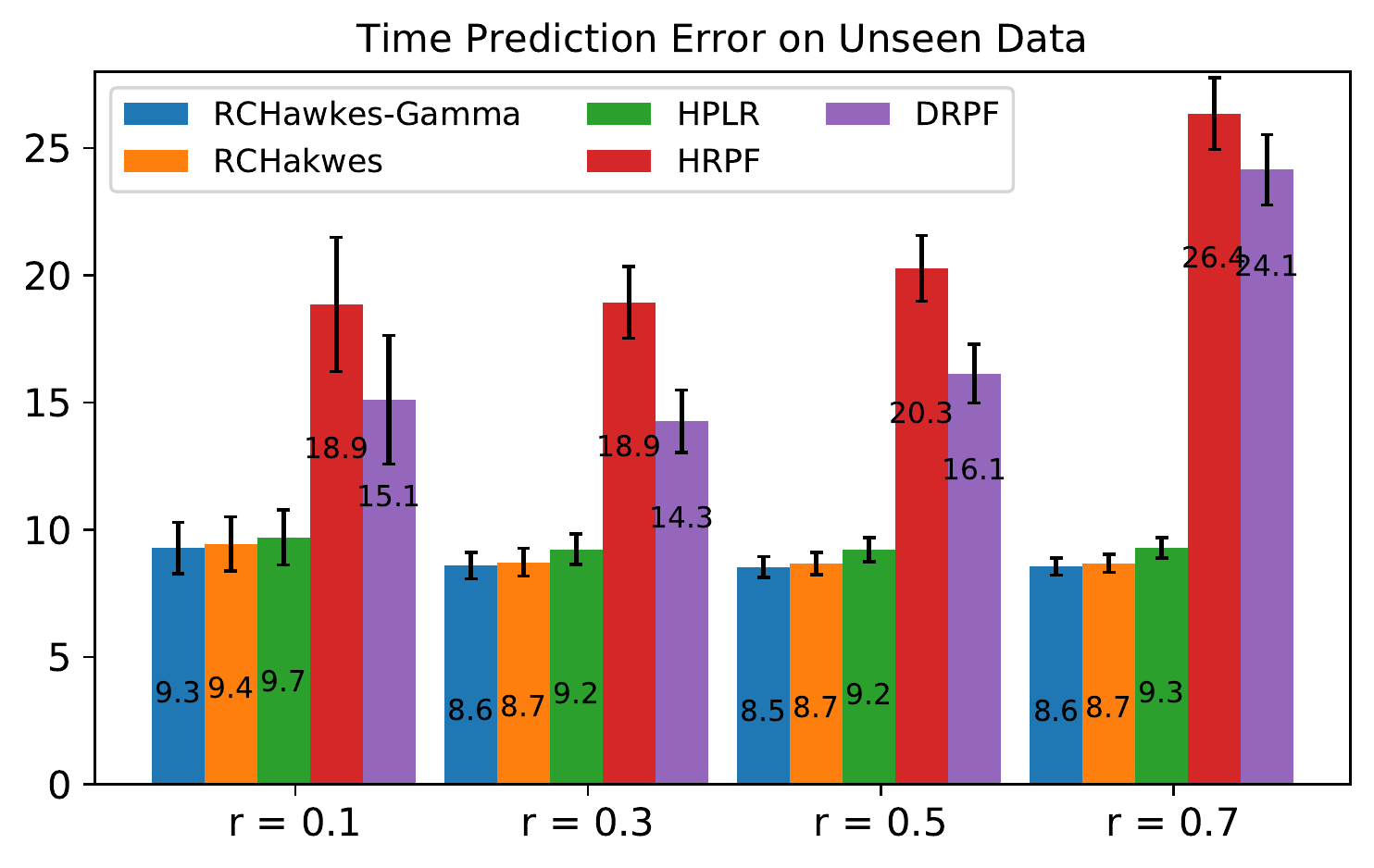}
}
\vspace{-15pt}
\caption{Time prediction error and 95\% confidence interval on synthetic datasets with varying data missing ratios ($r$)}
\label{fig:exp_res_syn}
\vspace{-15pt}
\end{figure*}

\begin{figure*}[!ht]

\centering

\subfigure{
\includegraphics[width=0.545\textwidth]{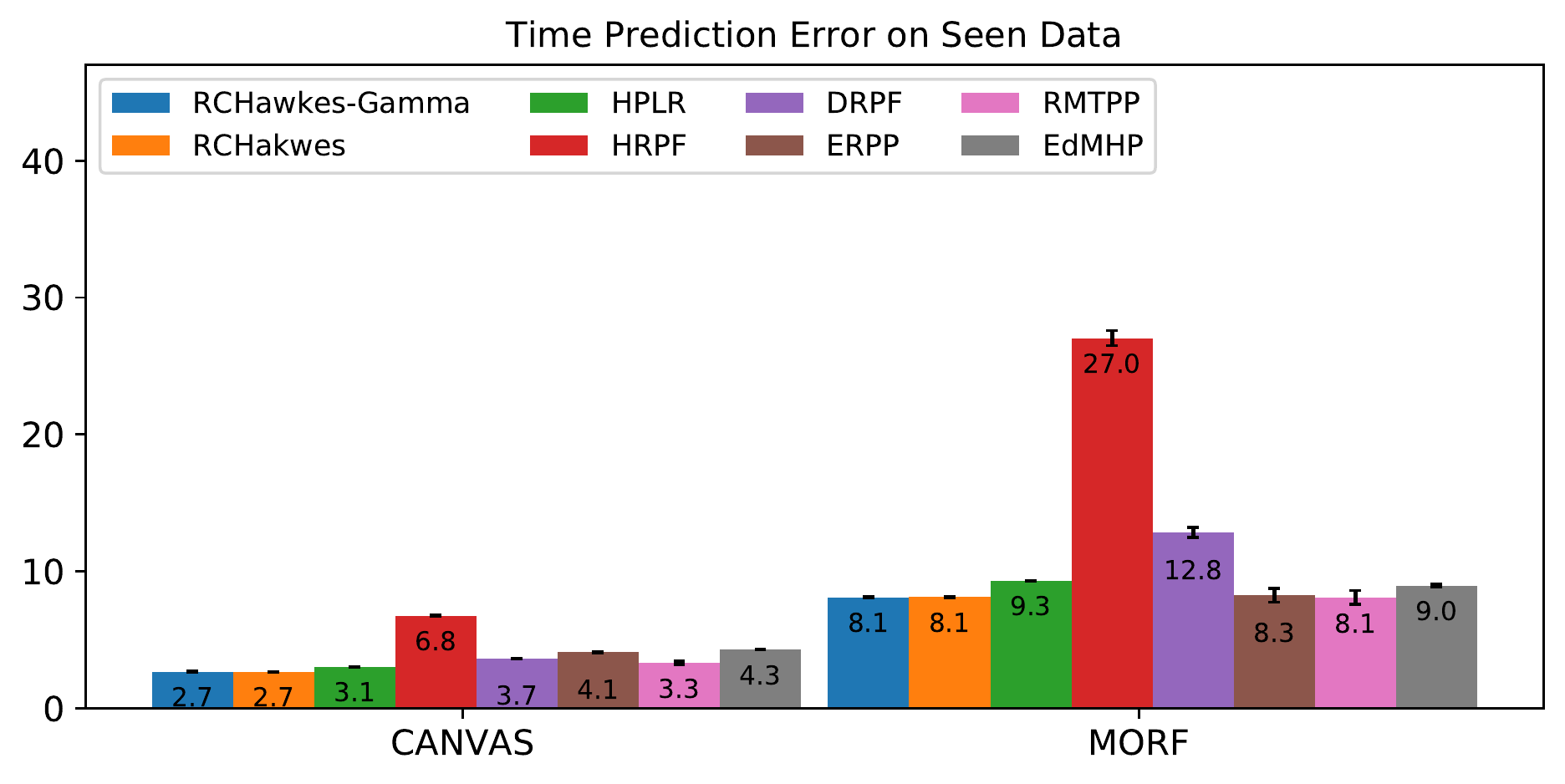}
}
 \hspace{-10pt}
\subfigure{
\includegraphics[width=0.435\textwidth]{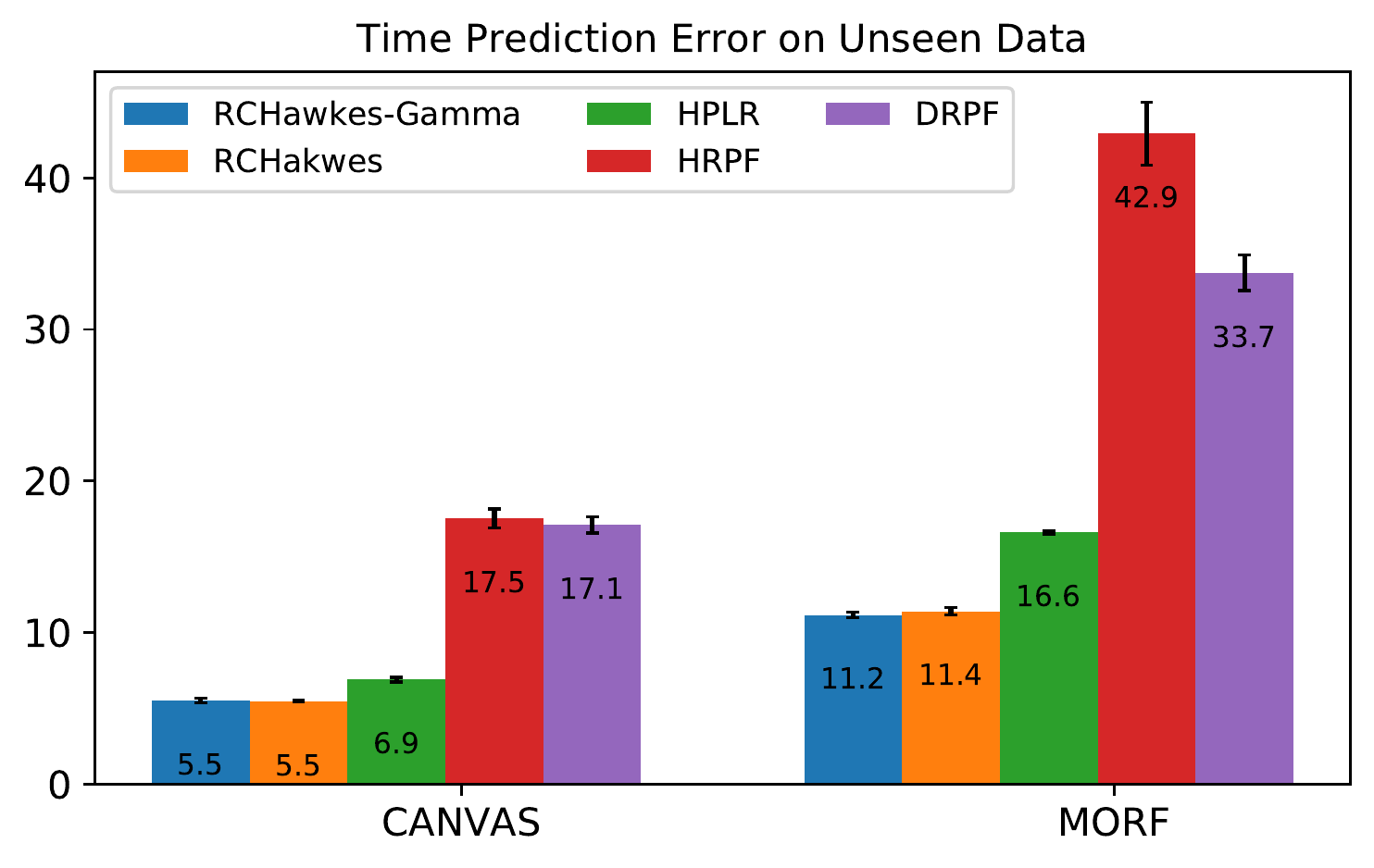}
}
\vspace{-15pt}
\caption{Time prediction error on seen and unseen data with 95\% confidence interval on real-world datasets }
\label{fig:exp_res_real}
\vspace{-15pt}
\end{figure*}

\subsubsection{Clustering Structure of Hawkes Parameters}
\label{sec:res:cluststr}
\begin{figure}[ht]
    \centering
    \includegraphics[width = \linewidth]{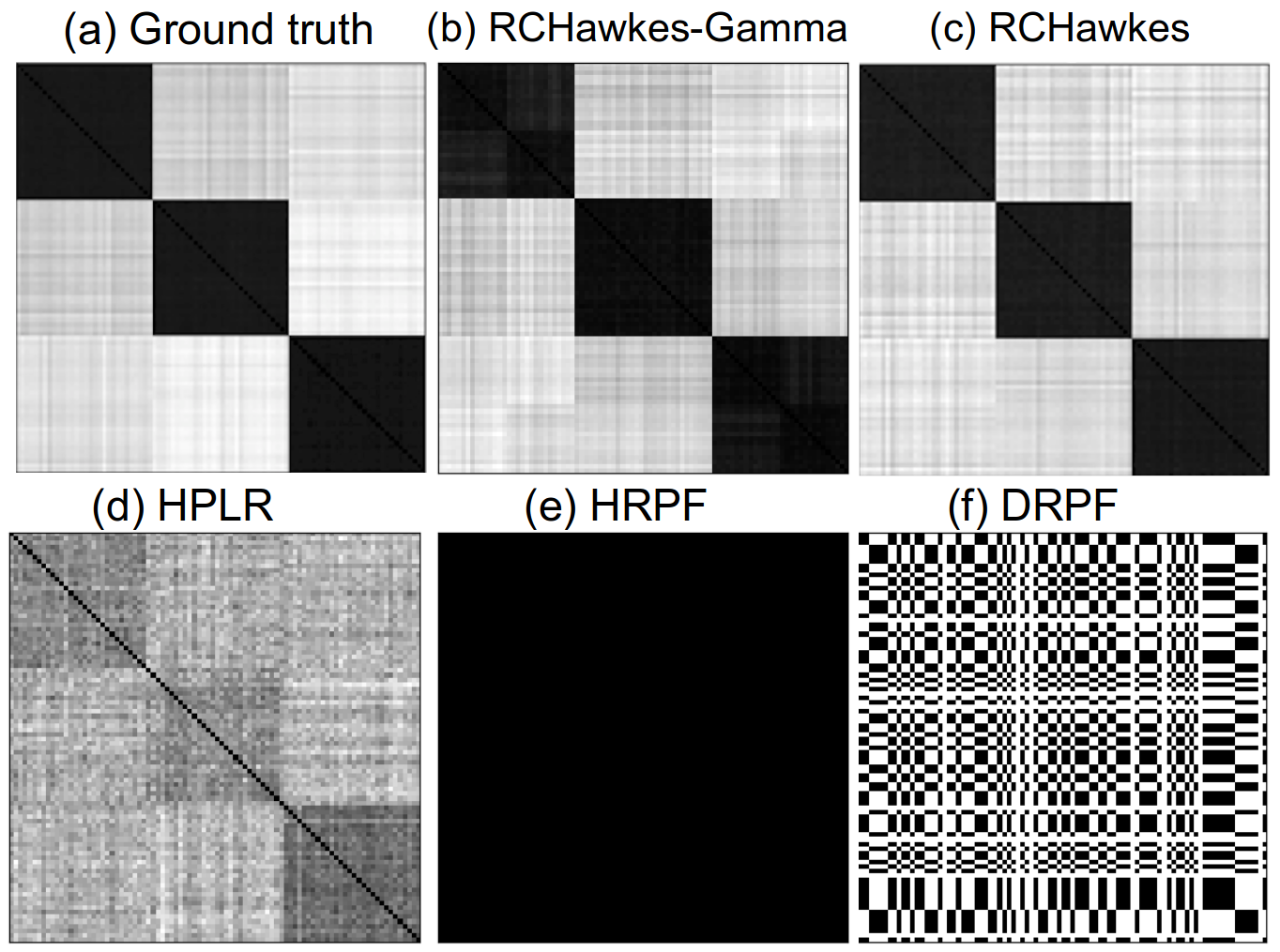}
    \caption{The ground truth of $A$'s correlation matrix (a), and the estimated $\hat{A}$'s correlation matrix learned by each model.}
    \label{fig:ep1}
    % \vspace{-15pt}
\end{figure}

% To see if the cluster structure of students is well captured by each model, we recover the order of the estimated parameter matrix $\hat{A}$'s columns learned by each model, so that all students who were originally in the same cluster are now grouped together. 
To see if the cluster structure of students is well captured by each model, we compute and present the correlation matrix of $\hat{A}$ between students with the recovered cluster orders in Figure~\ref{fig:ep1}. 
% Figure~\ref{fig:ep1} shows the plots of these correlation matrices in greyscale. 
% Specifically, each block in the correlation matrix represents the correlation matrix computed between students in one cluster to another (if not on the diagonal), or to themselves (if on the diagonal). 
% The darker the block, the higher the correlation. We also show the ground truth in Figure~\ref{fig:ep1} (a) as a reference. 
%As we can see, t
Our proposed models recover this structure closer to the ground truth (Figure~\ref{fig:ep1} (a)), i.e. a higher correlation within clusters (darker blocks) and a lower correlation between clusters (lighter blocks). 
HPLR introduces unnecessary correlations between clusters, possibly because of not having the student cluster assumption. % that students are not related. 
HRPF simply assumes all assignment-student pairs share the same parameter, thus has a meaningless correlation of 1 among all students. 
Finally, although DRPF improves HRPF by considering activity self-excitements, it fails to capture any meaningful correlation within clusters.
% \begin{figure}[ht]
% \vspace{-15pt}
% \centering
% \subfigure[Ground truth]{
% \includegraphics[width=0.145\textwidth]{figs/Ground_truth.png}
% }
%  \hspace{-12pt}
% \subfigure[\small{RCHawkes-Gamma}]{
% \includegraphics[width=0.15\textwidth]{figs/corr_matrix_CJHawkes_gamma.png}
% }
%  \hspace{-12pt}
% \subfigure[RCHawkes]{
% \includegraphics[width=0.15\textwidth]{figs/corr_matrix_CJHawkes.png}
% }
%  \hspace{-12pt}
% \subfigure[HPLR]{
% \includegraphics[width=0.15\textwidth]{figs/corr_matrix_lowrank.png}
% }
%  \hspace{-12pt}
% \subfigure[HRPF]{
% \includegraphics[width=0.15\textwidth]{figs/corr_matrix_HRPF.png}
% }
%  \hspace{-12pt}
% \subfigure[DRPF]{
% \includegraphics[width=0.15\textwidth]{figs/corr_matrix_DRPF.png}
% }
% \vspace{-15pt}
% \caption{The ground truth of $A$'s correlation matrix (a), and the estimated $\hat{A}$'s correlation matrix learned by each model.}
% \label{fig:ep1}
% \vspace{-20pt}
% \end{figure}

% 
\subsubsection{Returning Time Prediction on Simulated and Real Data}
\label{sec:res:rettime}

% \begin{figure*}[ht]

% \centering

% \subfigure{
% \includegraphics[width=0.545\textwidth]{figs/Sythetic_seen.pdf}
% }
%  \hspace{-10pt}
% \subfigure{
% \includegraphics[width=0.435\textwidth]{figs/Sythetic_unseen.pdf}
% }
% \vspace{-15pt}
% \caption{Time prediction error and 95\% confidence interval on synthetic datasets with varying data missing ratios ($r$)}
% \label{fig:exp_res_syn}
% \vspace{-15pt}
% \end{figure*}

% \begin{figure*}[!ht]

% \centering

% \subfigure{
% \includegraphics[width=0.545\textwidth]{figs/Real_seen.pdf}
% }
%  \hspace{-10pt}
% \subfigure{
% \includegraphics[width=0.435\textwidth]{figs/Real_unseen.pdf}
% }
% \vspace{-15pt}
% \caption{Time prediction error on seen and unseen data with 95\% confidence interval on real-world datasets }
% \label{fig:exp_res_real}
% \vspace{-15pt}
% \end{figure*}

In these experiments, we use a popular metric that has been used in many Hawkes-based models, i.e. RMSE on Time Prediction (TP), where TP is defined on the estimated next activity time, given the observed history (e.g. \cite{Du2013}). 
The baselines that do not directly infer parameters on unseen data (the future assignment scenario) are not included in the unseen data evaluation. 
Following the method used in~\cite{Du2013}, we sampled future activities based on the learned parameters via Ogata's thinning algorithm.

Figures~\ref{fig:exp_res_syn} and~\ref{fig:exp_res_real} respectively present the prediction error and $95\%$ confidence interval on simulated and real-world data.
As we can see from Figure~\ref{fig:exp_res_syn}, the proposed methods RCHawkes-Gamma and RCHakwes consistently outperform other baselines in all settings, except when the missing ratio is 0.1.
In that case, RMTPP and ERPP achieve the smallest error in seen data. 
However, unlike the proposed models that are almost invariant to the increase of missing ratio, ERPP and RMTPP's performances change dramatically with increasing $r$. 
More importantly, they lack the ability to directly predict the next activity time when the activity history is unseen. 
% We can also see that HRLR has comparable performance to the proposed method in synthetic data. 
% On the other hand, HRPF and DRPF have the highest prediction errors, possibly because they do not directly model self-excitement of activities. 
% EdMHP and RMTPP, even though they do not have the worst performance, lack the ability to directly predict the next activity time when the activity history is unseen. 
When comparing the baselines in the real datasets (Figure~\ref{fig:exp_res_real}), all approaches perform better in CANVAS dataset, compared to MORF. 
One possible explanation is that, in CANVAS, each assignment-student pair contains more historical activity time stamps. 
Therefore it provides all approaches more training data than the MORF dataset. 
% Furthermore, it shows that the proposed methods RCHawkes-Gamma and RCHakwes achieve significantly better performances compared with other baselines.
%especially, the performances of all approaches are very similar comparing to simulated data. 
% \textcolor{blue}{Despite that HPLR usually achieves lower predictive error than ERPP in synthetic data, ERPP works generally better than HPLR in real datasets.}

%% file: analysis.tex
Our application goal is to study students' cramming behaviors in MOOCs by modeling students' historical learning activities. 
% In the previous section, we have shown that the proposed RCHawkes-Gamma method has a better performance in modeling students' activity trajectories compared to the baseline methods.
In the following section, we will switch our focus to finding the connections between the characteristics of students' learning activities (parameterized by our model) and students' cramming behaviors. 

\subsubsection{Students' Cramming Behaviors}
\label{sec:res:cramming}
Since procrastination does not have a quantitative definition, in the first step of our analysis, we define the following measure to describe the degree of student procrastination presented in MOOCs: 
% if we denote the earliest possible time for a student $j$ to start assignment $i$ as $t^s_{ij}$ (start time), and use $t^a_{ij}$ to denote time when $j$'s first activity towards assignment $i$ takes place (action time), and let $t^d_{ij}$ to represent the latest time for student $i$ to work on assignment $i$ (end time), then we use 
$\text{delay} = \frac{t^a_{ij} - t^s_{ij}}{t^d_{ij} - t^s_{ij}}$ to quantify student $j$'s normalized delay in starting any activity that is associated with assignment $i$, where superscript $s,a,d$ respectively represents the start of the assignment, the first, and the last activity in the student-assignment pair.
Intuitively, this measure is the absolute time that student $j$ delays in starting assignment $i$, normalized by the duration that assignment $i$ is available for student $j$. 
Note that this measure is just a simple representation and cannot replace our model in predicting next activity times or uncovering cluster structures.
% In the MORF online dataset, the association information between assignments and student activities
%on video lectures and discussions 
% are presented.
% Consequently, we set the earliest activity that is associated with assignment $i$ %(could be viewing video or posting a question on discussion forum) 
% as the action time $t^a_{ij}$ and the end time $t^e_{ij}$ is set to be the deadline of assignment $i$. For the start time, if assignment $i$ is published after the first lecture that is associated with this assignment we set it to $t^s_{ij}$; otherwise, we set it to the publish time of the assignment. 
% In CANVAS, we set the deadline of assignment $i$ as the end time $t^e_{ij}$ and its publish time as start time $t^s_{ij}$. But, because of the nature of this dataset %as we do not know which module or discussion activities are associated with assignments, 
% we empirically set the first activity occurred after $t^s_{ij}$ as the action time $t^a_{ij}$. 

\subsubsection{Correlation Analysis}
\label{sec:res:corran}
In order to show how students activities parameterized by Hawkes and student delays are associated, we compute the Spearman's rank correlation coefficient between each pair of the variables. 
\begin{table}[h]
% \vspace{-7pt}
   \caption{Spearman's correlation between learned parameters and computed normalized student delays. p$<$0.001\text{***}  p$<$0.01  \text{**} p$<$0.05\text{*}}
   \vspace{-10pt}
      \centering        
    %   \footnotesize
    \resizebox{0.35\textwidth}{!}{
        \begin{tabular}{|c|c|c|c|c|}
        \hline
    \multicolumn{2}{|c|}{}               & $\alpha$                         & $\mu$                            & delay \\ \hline
    \multirow{3}{*}{CANVAS}
     &$\alpha$ & 1 & &  \\\cline{2-5}
      &  $\mu$ & 0.284\text{***}& 1 & \\\cline{2-5}
      & delay & 0.345\text{***} & 0.144\text{***}&1\\\hline

    \multirow{3}{*}{MORF}
        &$\alpha$ & 1 & &  \\\cline{2-5}
        &$\mu$ & 0.243\text{***}& 1 & \\\cline{2-5}
        &delay & 0.264\text{***} & 0.412\text{***}&1\\
\hline
    \end{tabular}}
% \caption{Spearman's correlation between learned parameters and computed normalized student delays. p$<$0.001\text{***}  p$<$0.01  \text{**} p$<$0.05\text{*} p$<$0.1$^{\cdot}$}
\label{tbl:corr}
\vspace{-5pt}
\end{table}
We choose the Spearman's correlation because it does not assume a normal distribution on the parameters, nor a linear relationship between the variables as Pearson correlation does. 
As we can see in Table~\ref{tbl:corr}, the two-sided p-values suggest that the correlations between these variables are statistically significant. We can also see that all the correlation coefficients are positive, meaning that student delays are positively associated with the base rate, i.e. expected number of occurrences per unit time that are excited by external stimuli (for example deadlines), and the burstiness of the occurrences. 
On the other hand, by looking at the two courses side-by-side, we can see that delay is more strongly associated with $\alpha$ in CANVAS. 
But, its association with the base rate $\mu$ is stronger in MORF. 
This suggests two different kinds of relationships between students and assignments: while in CANVAS big bursts of activities might suggest delays, in MORF small but frequent activities are associated with student delays.
%Note that our model goes beyond the association with delay. 
% and predicts the next activity times and recovers unseen cluster structures.
\subsubsection{Clustering Analysis}
\label{sec:res:clproc}
% As one of the assumptions in proposed model is the cluster structure among students in terms of self-excitement matrix $A$, we want to examine if this assumption is meaningful in real data.
% To examine our assumption of cluster structures among student cramming behaviors, we experiment to see if there are meaningful differences in the delay measure for students in different clusters.
To see if there are meaningful differences in the delay measure for students in different clusters, we first cluster the students using the K-means clustering algorithm, which has a similar objective to the cluster-structure term in our model (Eq.~\ref{eq:kmeans_org}), on the learned $\hat{A}$ matrix. 
%We choose this algorithm since the regularization term that imposes our cluster structure assumption makes use of the K-means objective function. 
Specifically, student $u_j$ is represented by the vector of estimated self-excitement parameters ($\hat{\alpha}_{1,i},...,\hat{\alpha}_{N,i}$) that are learned by RCHawkes-Gamma, and the cluster number for K-means is decided via grid search by looking at SSE. 
%We then examine student delays within each cluster for each assignment. 
%Finally, 
To examine the possible differences between clusters of students in terms of student delays, we conduct the Kruskal-Wallis test on all student delays across the clusters for each assignment. 
We report the average delay of all students in each cluster and for each assignment. 
The results are shown in Table~\ref{tab:kruskal_canvas} for CANVAS and in Table~\ref{tab:kruskal_coursera} for MORF dataset.
\begin{table}[]
    \centering
    % \footnotesize
    \caption{Kruskal Wallis test on delays in different clusters in CANVAS dataset. p$<$0.001\text{***}  p$<$0.01  \text{**} p$<$0.05\text{*}}  
    % \vspace{-10pt}
    \resizebox{0.48\textwidth}{!}{
    \begin{tabular}{|c|c|c|c|c|c|}
    \hline
    Assign. \#. & cluster 1 & cluster 2& cluster 3& cluster 4&p-value\\\hline
    size & 81 & 144 & 207 & 39 & - \\\hline
1&0.3335&	0.4583&	0.6108&	0.9064&	1.34E-16\text{***}\\
2&0.6245&	0.5788&	0.8476&	1.0854&	3.59E-09\text{***}\\
3&0.6911&	0.7143&	0.8633&	0.9655&	4.36E-05\text{***}\\
4&0.6050&	0.6958	&0.8515&	1.0717&	0.0008\text{***}\\
5&0.5969&	0.7080&	0.9084&	1.1217&	0.0195\text{*}\\
6&0.5351&	0.7647&	0.9002&	1.0970&	0.0149\text{*}\\\hline
    \end{tabular}}
    % \caption{Kruskal Wallis test on delays in different clusters in CANVAS dataset. p$<$0.001\text{***}  p$<$0.01  \text{**} p$<$0.05\text{*} p$<$0.1$^{\cdot}$}
    \label{tab:kruskal_canvas}
    % \vspace{-12pt}
\end{table}
\begin{table}[t!]
    \centering
    % \footnotesize
        \caption{Kruskal Wallis test on delays in different clusters in MORF dataset. p$<$0.001\text{***}  p$<$0.01  \text{**} p$<$0.05\text{*}}
           \vspace{-10pt}
    \resizebox{0.43\textwidth}{!}{
    \begin{tabular}{|c|c|c|c|c|}
    \hline
    Assign. \#. & cluster 1 & cluster 2& cluster 3&p-value\\\hline
    size & 573 & 34 & 68 & -\\\hline
1&0.4991&	0.6710&	0.4477&	2.30E-09\text{***}\\
2&0.5120&	0.7288&	0.4855&	1.90E-08\text{***}\\
3&0.5570&	0.6904&	0.6105&	7.50E-05\text{***}\\
4&0.4699&	0.6122&	0.5360&	0.0004\text{***}\\
5&0.5626&	0.6358&	0.6308&	0.0070\text{***}\\
6&0.5329&	0.6236&	0.6642&	8.56E-06\text{***}\\
7&0.4325&	0.5598&	0.7672&	2.12E-20\text{***}\\
8&0.3974&	0.5172&	0.7629&	3.84E-27\text{***}\\\hline
    \end{tabular}}
    % \caption{Kruskal Wallis test on delays in different clusters in MORF dataset. p$<$0.001\text{***}  p$<$0.01  \text{**} p$<$0.05\text{*} p$<$0.1$^{\cdot}$}
    \label{tab:kruskal_coursera}
% \vspace{-10pt}
\end{table}
In CANVAS, 4 student clusters are found. 
These clusters all have significant differences in terms of delays. 
For example, students in cluster 1 have the smallest delay, with a general decreasing trend towards the later assignments. 
On the other hand, delays are the worst for students in cluster 4, with an average delay greater or close to 1 for all assignments, which implies that this group of students tend to start the assignments very close to or even later than the deadline. 
In the 3 clusters that are found in the MORF dataset, the p-values of Kruskal-Wallis tests show strong evidence of cluster differences for each assignment. 
Specifically, the majority of the students in the MORF course are in cluster 1 and their delays are overall the lowest comparing to the other two clusters. 
They tend to delay less and less over time. 
On the other hand, students in cluster 3 start the course with a low delay but increase their delay so fast that at the end of the course, they turn out to be the students who delay the most. 
This analysis demonstrates that the self-excitement parameters have strong associations with student delays, which not only reinforces the findings from the correlation analysis, but also suggests that they are good indicators in characterizing students' cramming behaviors.

%% file: conclusion.tex
 In this paper, we proposed a novel uni-variate clustered Hawkes process model, RCHawkes-Gamma to model procrastination behaviors of students in MOOCs. 
 Particularly, the proposed method models activities on all assignment-student pairs jointly and assumes cluster structures between students and relatedness between assignments.
 We test our proposed model on a synthetic dataset and two real-world MOOC datasets. 
 The results of our experiments show that our proposed model can predict students' next activity time with lower time prediction error on both seen and unseen data, compared to the baseline methods.
 We also study and analyze the parameters learned by the proposed model on both MOOC datasets.  
 Our analysis reveals the positive associations between student delays with our model's parameters.
 The model also discovers meaningful clusters of students who show different delaying behavior trends. 
 It is also worth noting that our proposed approach can be useful in real-world scenarios such as for professional educators or adaptive learning systems. 
 As an example, the prediction of future activities especially on the unseen student-assignment pairs can provide teachers the opportunity to intervene with students who show strong procrastination tendencies. For students, their learning activities can be presented in formats such as a dashboard, for visualization, summarization, and feedback generation, which in turn can be beneficial in regularizing students' learning behaviors.
While our model is created with the education domain in mind, it can be applied to other domains such as recommender systems. 
% For example, by modeling user-item pairs (i.e. users buying histories for items),
% our model can be used to predict users' potential return time to items in the future.
 %---------------------------------------- 
 
 A limitation of this work is that the delay measure is used as a proxy for procrastination, while self-reported procrastination measures could have helped in labeling delays more accurately as procrastination. %can be more precise if available.
%  Furthermore, our model treats all activities as one type, and assumes that the same parameters are shared across all different types, which is not necessarily true in many other applications.
 Furthermore, our proposed method does not aim to differentiate between active procrastination (i.e. due to the internal need to experience thrill by delaying the tasks to the last minute) and passive procrastination (irrational delay despite expecting negative results due to the delay), as indications such as purposeful, strategic, and arousal delay can not be inferred from the datasets used in this work.
 %---------------------------------------- importance of findings

%% file: algorithm.tex
\section{Appendix}
\subsection{Algorithm 1}
\subsubsection{Algorithm Walk-through}
In the following, we provide some details of Algorithm 1 shown as below.
Specifically, the the following subroutine is repeated in the algorithm:
\input{algo}
\noindent
(1) Computation of $A_*$ (lines 8-9): The objective of this part is defined as follow:
% \begin{align}
% \label{eq:proximal}
%      &\min_{A_z,U_z,Z_z} \|A_z-A_s\|_F^2 + \|U_z-U_s\|_F^2 + \|Z_z-Z_s\|_F^2 + \\\nonumber
% & \text{s.t. } tr(Z_z) = k, tr(A_z) \leq c, A_z\geq0, U_z\geq0, Z_z\preceq I, Z_z\in S_{+}^M\nonumber 
% % \label{eq:proximal}
% \end{align} 
\begin{align}
\label{eq:A}
    \min_{A_z} F_A (A_z) :=\|A_z-A_s\|_F^2~\text{s.t.}~tr(A_z) \leq c, A_z\geq0.
\end{align}
by following the Accelerated Gradient Method schema, we compute $A_* = \mathcal{M_{\gamma,S^A}}$ (line 8), where $\mathcal{M_{\gamma,S^A}} := \frac{1}{\gamma}\|A - \big (S^A-\frac{1}{\gamma}\nabla \mathcal{L}(A)\big )_+\|_F^2 + \rho_3 tr(A)$~\cite{ji2009accelerated}; where $S^A$ is current search point; $\gamma$ is the step size; and $\rho_3$ is the regularization coefficient. Specifically, we use trace norm projection (TrPro)~\cite{cai2010singular} to solve the above minimization problem. Finally $(\cdot)_+$ projects negative values to 0 as we constraint $A$ to be nonnegative. 
% Similarly, we can compute $U_{*}$

\noindent
(2) Computation of $U_*$ (line 10): similarly to the compuation of $A_*$, we compute optimal value of $U$, $U_*  = \mathcal{M}_{S^U_i,\gamma_i}(U)$, where $S^U$ is the current search point of $U$, and $(\cdot)_{+}$ is the nonnegative projection. Specifically the objective of this computation is:
\begin{align}
\label{eq:U}
    \min_{U_z}F_U(U_z):= \|U_z-U_s\|_F^2~\text{s.t.} ~ U_z\geq0.
\end{align}

\noindent
(3) Computation of $Z_*$ (lines 11-14): as the constraints on $Z$ are more complicated, the proximal operator also has more terms. Specifically, the goal is to solve the following optimization problem:
\begin{align}
% \label{eq:Z}
    \label{eq:m_opt}
    \min_{Z_z} \|Z_z - \hat{Z_s}\|_F^2, \text{ s.t. } tr(Z_z) = k,Z_z\preceq I, Z_z\in S_{+}^M
    % \label{eq:m_opt}
\end{align}

To solve this problem, we apply eigen decomposition on $Z_i$ such that $Z_i = Q\Sigma Q'$, where $\Sigma = diag(\hat{\sigma}_1,...,\hat{\sigma}_M)$. 
It has been shown that $Z_* = Q\Sigma_{*}Q'$, where $\Sigma_{*} = diag(\sigma^*_1,...,\sigma^*_k)$, and $\sigma^*_i$ is the optimal solution to the problem~\cite{zha2002spectral}:
\begin{align}
% \label{eq:Z-alt}
    \label{eq:sigma_opt}
    \min_{\Sigma} \|\Sigma_* - \Sigma\|_F^2, \text{ s.t. } \sum_i^M \sigma_i  = k,~0\leq \sigma_i\leq 1.
\end{align}
To solve Eq.~\ref{eq:sigma_opt} with constraints, we apply the linear algorithm proposed in~\cite{kiwiel2007linear}. 

Remark: we want to quickly show that by solving problem~\ref{eq:sigma_opt}, the resulting $Q\Sigma_*Q'$ provides a closed-form solution to Eq.~\ref{eq:m_opt}. If denote eigen-decomposition of $M_z = P\Lambda P'$, by definition, $P'P = PP' = I$ and $\Lambda = diag(\lambda_1,...,\lambda_M)$ where $\lambda_i$ for $i = 1,...,M$ are eigenvalues of $z$. Then Eq.~\ref{eq:m_opt} can be equivalently written as:
\begin{align}
\label{eq:rewrite_m}
    &\min_{\Lambda,P}\|Q'P\Lambda P'Q - \Sigma\|_F^2~\text{s.t.}~ tr(\Lambda) = k\\\nonumber
    &\lambda  = diag(\lambda_1,...,\lambda_M), 0\leq\lambda_i\leq1,P'P = PP' = I.
\end{align} 
It is easy to see that the constraints of the two equations with respect to $\Lambda$ and $\Sigma$ are equivalent. Furthermore, if denote the objectives of Eq.~\ref{eq:sigma_opt} and Eq.~\ref{eq:rewrite_m} as $f(\cdot)$ and $g(\cdot)$ respectively, by definition, the feasible domain of Eq.~\ref{eq:sigma_opt} is a subset of the feasible domain of Eq.~\ref{eq:rewrite_m}, therefore $f(\Sigma_*)\geq g(Q'P_*\lambda_* P_*'Q)$. On the other hand, knowing that $\Sigma$ is a diagonal matrix, $\|Q'P_*\lambda_* P_*'Q - \Sigma\|_F^2 \geq \|(Q'P_*\lambda_* P_*'Q)\circ I - \Sigma\|_F^2$, meaning that the optimal objective value of Eq.~\ref{eq:sigma_opt} is no greater than the optimal objective value of Eq.~\ref{eq:rewrite_m}. Therefore, the two problems are equivalent. 

\subsubsection{Complexity Analysis}
Recall that we consider the setting where there are $M$ students and $N$ assignments.
The complexity of the computation of $A_*$ (line $8-9$) is $\mathcal{O}(MN^2)$ where a truncate SVD is used. To solve Eq.~\ref{eq:m_opt}, we first apply eigen-decomposition on the $M\times M$ matrix $S_i^Z$ (in line 11), which has time complexity of $\mathcal{O}(M^3)$, then we solve Eq.~\ref{eq:sigma_opt} which has shown to be the closed-form solution to Eq.~\ref{eq:m_opt} (line 12), a complexity of $\mathcal{O}(M)$ can be achieved~\cite{kiwiel2007linear}. As we introduce recursive function $R$ in Sec. 4.4, the complexity of computing loss $\mathcal{L}$ (line $15$) is $\mathcal{O}(MNK)$ if let $K$ denote the number of activities of the longest student-assignment pair. Each line of the other parts of the algorithm requires $\mathcal{O}(MN)$ as only basic operations are involved. As a result, the time complexity per time step is $\mathcal{O}(\max(M,N)^2M + MNK)$. In the cases where conventional Hawkes model is used, without the help of recursive function $R$, computing the loss per time step needs $\mathcal{O}(MNK^2)$. 
Note that without operations such as truncated SVD, even though a complexity of $\mathcal{O}(MN^2)$ can be avoided for conventional Hawkes models, the parameters of student-assignment pairs that do not have observed activities can not be inferred.

When it comes to the number of parameters to be learned, for our model, due to our low rank and cluster structure assumption on $A\in\mathbb{R}^{N \times M}$, the number of parameters it requires to meet these two assumptions is $(M+N)c + 2Mk$ where $c<\min(M,N)$ and $k<M$ is respectively the rank of $A$ and the number of clusters among students, i.e. the rank of $Z\in\mathbb{R}^{M\times M}$. 
For conventional Hawkes models, each student-assignment pair needs to be learned independently. As a result, the number of parameters need to complete matrix $A$ is $M\times N$.

\subsubsection{Convergence Analysis}
As mentioned earlier in this section, we have shown that Algorithm 1 repeatedly solves the subroutines respectively defined in Eq.~\ref{eq:A},~\ref{eq:U} and~\ref{eq:m_opt}, where solving Eq.~\ref{eq:m_opt} is mathematically equivalent to solving Eq.~\ref{eq:sigma_opt}. 
As it is known that accelerated gradient descent can achieve the optimal convergence rate of $\mathcal{O}(1/k^2)$ when the objective function is smooth, and only the subrountine of solving Eq.~\ref{eq:A} involves non-smooth trace norm, the focus of the following section is to provide a convergence analysis on this subrountine.
% However, it is also shown in a more recent work of Nesterov et al. that, the same convergence rate can be achieved even when the objective is not smooth\cite{nesterov2013gradient}.
% In the following we provide the convergence analysis for each part of the subroutines. 
% \paragraph{Convergence of $F_A$.} 
% It is known that accelerated gradient descent can achieve the optimal convergence rate of $\mathcal{O}(1/i^2)$ when the objective function is smooth. However, it is also shown in a more recent work of Nesterov et al. that, the same convergence rate can be achieved even when the objective is not smooth\cite{nesterov2013gradient}. 
Specifically, by following the outline of proof provided in Ji and Ye's work~\cite{ji2009accelerated}, we show that a rate of $\mathcal{O}(1/\epsilon^2)$ can be achieved in solving Eq.~\ref{eq:A}, even with the presence of trace norm in the objective. Specifically, if let $A_*$ denotes the optimal solution, by applying Lemma 3.1 from Ji and Ye's work, we can obtain the following:
\begin{align}
\label{eq:FA-1}
    \medmath{F_A(A_*) - F_A(A_1)} & \medmath{\geq \frac{\gamma_1}{2}\|A_1 - S_1^A\|^2 + \gamma_1\langle S_1^A - A_*, A_1 - S^A_1\rangle}\\\nonumber
    &=\medmath{\frac{\gamma_1}{2}\|A_1 - A_*\|^2 - \frac{\gamma_1}{2}\|S^A_1 - A_*\|^2},
\end{align}
which is equivalent to:
\begin{align}
\label{eq:FA-2}
    \medmath{\frac{2}{\gamma_1}(F_A(A_1) - F_A(A_*))\leq \|S^A_1 - A_*\|^2 - \|A_1 - A_*\|^2.}
\end{align}
Then by following the proof of Theorem 4 in Ji and Ye's work, we can obtain the following inequality, using the equality $\alpha_i^2 = \alpha_{i+1}^2 - \alpha_{i+1}$ derived from the equation in line $24$ of our algorithm and the definition of $S^A_i$ in line 4:
\begin{align}
    \label{eq:FA-3}
& \medmath{\frac{2}{\gamma_{i+1}}\big[\alpha_i^2 (F_A(A_i) - F_A(A_*)) - \alpha_{i+1}^2\big((F_A(A_{i+1}) - F_A(A_*))\big)\big]} \\\nonumber
    &\medmath{\geq \|\alpha_{i+1}A_{i+1} - (\alpha_{i+1})A_i - A_*\|^2- \|\alpha_{i}A_{i}}\\\nonumber
   & \medmath{- (\alpha_{i}-1)A_{i-1} - A_*\|^2}.
\end{align}

As $\eta\geq 1$ and we update $\gamma_{i+1}$ by multiplying $\eta$ with $\gamma_i$, we know that $\gamma_{i+1}\geq \gamma_i$. By plugging in this inequality to Eq.~\ref{eq:FA-3}, we can obtain the following:
\begin{align}
\label{eq:FA-4}
   & \medmath{\frac{2}{\gamma_i}\alpha_i^2 (F_A(A_i) - F_A(A_*)) - \frac{2}{\gamma_{i+1}}\alpha_{i+1}^2 (F_A(A_{i+1}) - F_A(A_*))}\\\nonumber
   &\medmath{\geq \|\alpha_{i+1}A_{i+1} - (\alpha_{i+1}-1)A_i - A_*\|^2 - \|\alpha_{i}A_{i}}\\\nonumber
   &\medmath{- (\alpha_{i}-1)A_{i-1} - A_*\|^2.}
\end{align}

By summing up each side of Eq.~\ref{eq:FA-4} from $i  =1$ to $i=k$, then combining with Eq.~\ref{eq:FA-2}, we can obtain the following:
\begin{align}
    &\medmath{\frac{2}{\gamma_i}\alpha_{i}^2\big(F_A(A_i) - F_A(A_*)\big)\leq \|A_1 - A_*\|^2 }\\\nonumber
    & \medmath{- \|\alpha_i A_i - (\alpha_i -1 )A_{i-1} - A_*\|^2  
    + \frac{2}{\gamma_1} \big(F_A(A_1) - F_A(A_*)\big)}\\\nonumber
   & \medmath{\leq \|A_i - A_*\|^2 - \|\alpha_i A_i - (\alpha_i -1 )A_{i-1} - A_*\|^2  }\\\nonumber
   &\medmath{+ \|A_0 - A_*\|^2 - \|A_i - A_*\|^2 }\\\nonumber
   & \medmath{\leq \|A_0 - A_*\|^2}
\end{align}
Using the fact that $\alpha_i\geq \frac{i+1}{2}$ (can be shown using induction from line $24$ of the algorithm), we can obtain: 
\begin{align}
\medmath{
    F_A(A_i) - F_A(A_*) \leq \frac{2\gamma_i\|A_* - A_0\|^2}{(i+1)^2}.}
\end{align}

%% file: intuition.tex
\subsection{Intuition Explained}
Since our goal is to study sequences of student activities and their inter-arrival times, point processes are of the best choices for our application.
Poisson process assumes that the past and future activities are completely independent. Unlike the memory less nature of the Poisson process, the Hawkes process expects that activities to be exciting both externally (similar to the Poisson process) and internally, that is, activities are self-exciting.

From the branching process point of view of the Hawkes process, activities are assumed to have latent or unobserved branching structures, where the offspring activities (i.e. future activities) are triggered by parent activities (i.e. past activities) while the immigrant activities arrive independently.
Therefore, the offspring are also said to be structured into clusters.
In the online learning setting, smaller activity chunks towards a goal or deadline can be examples of offspring: students divide the big tasks (the whole process) into small sub-tasks (offspring clusters).
The deadline (external stimuli) of a big task (such as a task) triggers the follow-up activities related to small tasks, which come one after another in a so-called burst mode (self-excitement). 

\begin{figure}[ht!]
    \centering
    \includegraphics[width = 0.9\linewidth]{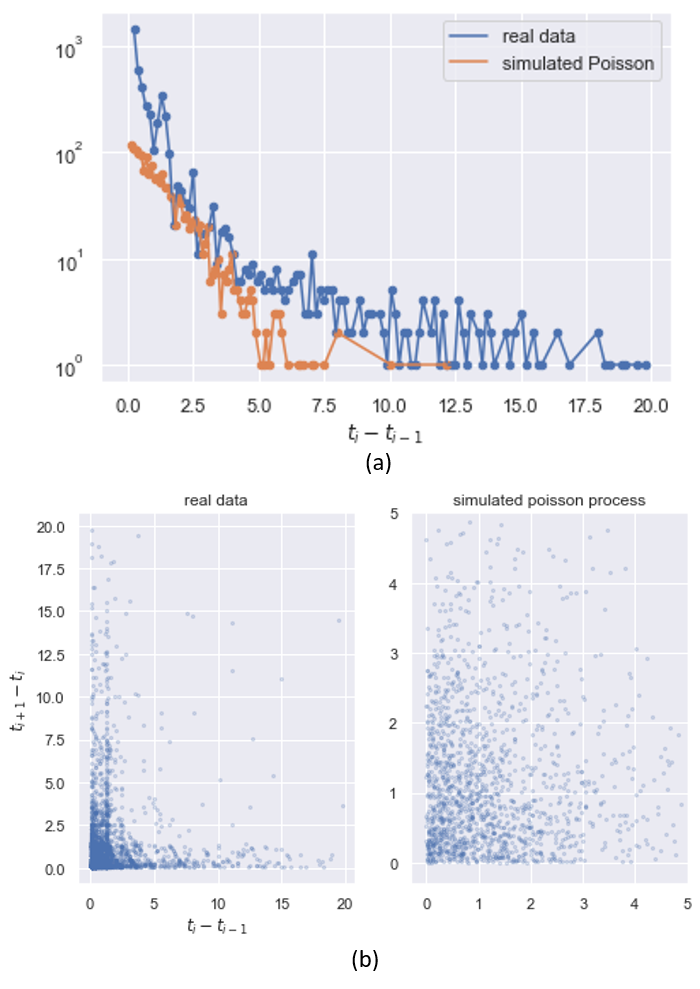}
    \caption{Two tests show the differences between a simulated Poisson process vs. a true process extracted from CANVAS dataset in terms of (a) inter-arrival times distributions and (b) inter-arrival times autocorrelation.}
    \label{fig:nonpoisson}
\end{figure}
To empirically demonstrate that self-excitement or burstiness is observed in the online course setting, we conducted two tests to show that Poissonian properties are not present in true student activity sequences. 
The first test is to check the distribution of the inter-arrival times, which is defined as the difference between two consecutive activity occurrences' arrival times. 
In Figure.~\ref{fig:nonpoisson} (a), we show the inter-arrival times versus simulated Poisson process in a real student's sequence of activities for an assignment. 
The simulated Poisson process is generated with the same average rate, as the real student's sequence, on a log-log scale. 
We see that the Poisson process almost forms a straight line, indicating the exponential distribution of inter-arrival times, whereas the real data is ``nonpoissonian'', i.e. includes short pauses followed by long ones. 
The second test is to check the 1-lag autocorrelation of inter-arrival times. 
As we can see in Figure.~\ref{fig:nonpoisson} (b), no autocorrelation is spotted in the Poisson process, whereas the real data exhibits some pattern: dense activities followed by long pauses.